*Research Article*

# Transferable Feature Representation for Visible-to-Infrared Cross-Dataset Human Action Recognition

**Yang Liu ,[1,2] Zhaoyang Lu,[1,2] Jing Li,[1,2] Chao Yao ,[3] and Yanzi Deng[1,2]**

[1]*School of Telecommunications Engineering, Xidian University, No. 2 South Taibai Road, Xi'an, Shaanxi 710071, China*
[2]*State Key Laboratory of Integrated Services Networks, Xidian University, No. 2 South Taibai Road, Xi'an, Shaanxi 710071, China*
[3]*School of Automation, Northwestern Polytechnical University, No. 127, West Youyi Road, Xi'an, Shaanxi 710072, China*

Correspondence should be addressed to Chao Yao; yaochao@nwpu.edu.cn

Recently, infrared human action recognition has attracted increasing attention for it has many advantages over visible light, that is, being robust to illumination change and shadows. However, the infrared action data is limited until now, which degrades the performance of infrared action recognition. Motivated by the idea of transfer learning, an infrared human action recognition framework using auxiliary data from visible light is proposed to solve the problem of limited infrared action data. In the proposed framework, we first construct a novel Cross-Dataset Feature Alignment and Generalization (CDFAG) framework to map the infrared data and visible light data into a common feature space, where Kernel Manifold Alignment (KEMA) and a dual aligned-to-generalized encoders (AGE) model are employed to represent the feature. Then, a support vector machine (SVM) is trained, using both the infrared data and visible light data, and can classify the features derived from infrared data. The proposed method is evaluated on InfAR, which is a publicly available infrared human action dataset. To build up auxiliary data, we set up a novel visible light action dataset XD145. Experimental results show that the proposed method can achieve state-of-the-art performance compared with several transfer learning and domain adaptation methods.

## 1. Introduction

Human action recognition aims to recognize an ongoing action from a video clip, which has received great attention in recent years due to its wide applications, including video surveillance [1], video labeling [2], video content retrieval [3], human-computer interaction [4], and sports video analysis [5]. Over the past decades, significant progress has been made in action recognition [6] and most of the state-of-the-art approaches for action recognition have been put into visible imaging videos [7–9]. In addition, many visible light action datasets have been constructed for action recognition, such as KTH [10], HMDB51 [11], and UCF101 [12]. Generally speaking, human action recognition in visible light has been well addressed and successfully applied to some applications. However, illumination change, shadow, background clutter, and occlusion remain to be great challenges for visible light action recognition [13].

With the development of sensor technology, human actions can be captured by the thermal infrared cameras instead of the visible light ones. Compared with visible light action recognition, infrared action recognition can solve the aforementioned challenges [14]. For example, the infrared thermal imaging is robust to illumination change because it can capture humans well under poor light condition when the person can almost not be seen in the visible light videos, which is very useful for night surveillance or human-computer interaction (HCI) under dim night. In addition, as the temperatures of the shadow, background clutter, and occlusion are relatively low compared with that of the humans or moving objects in infrared videos, these challenges can be well suppressed in infrared videos. With these properties, infrared action recognition can be adopted in more applications and outperform that in visible light. Therefore, infrared action recognition may become a next hot topic in computer vision in the future.

Actually, infrared and visible action data lie in different feature space and the traditional approaches for visible light action recognition cannot be directly applied to infrared action recognition due to the modality gap between them. However, the methods for infrared action recognition are limited. Furthermore, there is only one publicly available infrared dataset InfAR [20] for action recognition until now. As a result, the performance of infrared action recognition in previous works is preliminary and leaves a reasonable space to further promote its performance. To these issues, if a large amount of previously annotated videos from various visible light videos datasets can be transferred to infrared domains for recognition, considerable amount of time-consuming human efforts, such as collecting and hand labeling a large amount of infrared action videos, can be saved. In addition, as infrared and visible light videos may contain complementary information, infrared action recognition performance can be improved if the knowledge from visible light and infrared data can be properly integrated. Nevertheless, it would have at least two obstacles to integrate these two datasets. (1) Infrared and visible light videos are captured by different sensors; the strong modality gap between them will degrade the recognition performance without effective transferable feature representation. (2) In real-world scenario, infrared videos are usually limited while the visible ones are abundant. The imbalanced data distribution will also degrade the classification performance.

Tackling these problems, a novel Cross-Dataset Feature Alignment and Generalization (CDFAG) framework is proposed for infrared action recognition task in this paper. To be more specific, we focus on adapting, aligning, and generalizing representations from different domains to a single common feature space in order to bring the original target domain (infrared action data) and the auxiliary source domain (visible light action data) into the same feature space. And then we learn a unique classifier in that semantically meaningful aligned and generalized feature space across datasets. In this way, the modality gap between these two datasets is reduced. To better use the data in the generalized feature space, we adopt semisupervised technique so that both the labeled and unlabeled data are considered in our method. In more detail, Kernel Manifold Alignment (KEMA) [15] is adapted to cross-dataset action recognition to generate aligned representations and then cross-domain generalized features are learnt by training two novel aligned-to-generalized encoders (AGE) on the source and target datasets in parallel. To build up source domain data, we set up a new visible light action dataset XD145. Putting all the things together, we can summarize the main contributions of this paper as follows:

(i) We have proposed a novel Cross-Dataset Feature Alignment and Generalization (CDFAG) framework to address the visible-to-infrared action recognition problem. It can efficiently reduce the modality gap across datasets and generate aligned and generalized feature representations in a common space with low intraclass diversity and high interclass variance.

(ii) It is the first time Kernel Manifold Alignment (KEMA) is applied in infrared action recognition

field to generate aligned representations in a common latent space.

(iii) We have designed a novel aligned-to-generalized encoder (AGE) model to learn generalized feature representations after feature alignment by KEMA.

(iv) We achieved state-of-the-art results in visible-to-infrared action recognition compared with several transfer learning, domain adaptation, and deep learning based methods.

(v) Since there are a limited number of action videos from existing benchmark visible light datasets which share the same class label with that of the InfAR dataset, we have constructed a new visible light action dataset called XD145 to build up auxiliary source domain data. And this dataset could be further utilized as the benchmark visible light action dataset.

The rest of the paper is organized as follows. In Section 2, we review some background and related works. In Section 3, we explain details of our proposed method. Section 4 presents the experimental results of our proposed method on visible-to-infrared action recognition and cross-dataset action recognition, and finally Section 5 draws the conclusion and future research lines.

## 2. Related Work

In this section, we present the background and related works. We briefly review the concepts, methods in transfer learning and domain adaptation, their benefits in cross-domain action recognition, and the development status of infrared action recognition.

*2.1. Transfer Learning and Domain Adaptation.* The classical pattern recognition and machine learning tasks [21–23] mainly adopt a robust classifier learnt by annotated training data and assume the testing data and the training data belong to the same feature space or distribution. However, it is unrealistic in real-world applications because of the high price of manual labeling training samples and environmental restrictions. Therefore, sufficient training data that share the same feature space or distribution with the testing data cannot always be guaranteed even using some feature selection methods [24, 25] without considering the distribution gap. In this case, the potential discriminability of the trained model can be limited by the insufficient training data. Of the several schools of thought addressing this problem, two prominent ones are transfer learning [26] and domain adaptation [27]. In fact, transfer learning methods are closely related but not equivalent to domain adaptation. Transfer learning aims to transfer the knowledge from a source domain to the target domain while domain adaptation methods are essentially solving transfer learning problems. Surveys like [26] show that the type of knowledge being transferred can be roughly classified into four categories: (1) instance transfer, (2) feature-representation transfer, (3) parameter transfer, and (4) relational-knowledge transfer. Our proposed method falls into feature-representation transfer by adapting the representations from different domains to a single common latent space. The literature of feature-representation transfer can be roughly divided into three kinds of adaptation problems: supervised, unsupervised, and semisupervised adaptation problems, depending on the availability of labels in different domains.

Semisupervised domain adaptation has attracted much attention in recent years. For example, a reconstruction-based domain adaptation method called latent sparse domain transfer (LSDT) was proposed in [28] for visual categorization of heterogeneous data via subspace learning and sparse representation. A $l_{2,1}$-norm based discriminative robust kernel transfer learning (DKTL) method was proposed in [29] to address distribution mismatch problem of image classification across domains. Although the methods in [28, 29] achieve good performance in cross-domain image classification, to the best of our knowledge, a more challenging problem of visible-to-infrared cross-dataset action recognition was not studied. Extreme learning machine was used in [30] to address the visual knowledge adaptation problem for video event recognition and object recognition. It should be noted that the proposed method is based on feature-representation domain adaption, which is essentially different from the extreme learning machine methods, which are classifier-based domain adaptation approaches. Apart from the methods mentioned above, manifold alignment is an important kind of semisupervised domain adaptation methods, which concurrently matches the corresponding samples and preserves the geometry of each domain by graph Laplacian [31]. Actually, data manifolds alignment boils down to finding projections to a common latent space. Semisupervised method Kernel Manifold Alignment (KEMA) was proposed in [15] and has been successfully applied to multimodal visual object recognition [15], multisubject facial expression recognition [15], and multitemporal remote sensing image classification [32]. Nevertheless, Kernel Manifold Alignment (KEMA) method has not been applied in cross-domain human action recognition. Therefore, we have studied the effectiveness of the Kernel Manifold Alignment (KEMA) and adapted it to visible-to-infrared action recognition to obtain aligned feature representations across datasets. The main focus of our work is on the use of KEMA for feature alignment in visible-to-infrared action recognition which was not addressed in [15].

### 2.2. Cross-Domain Action Recognition via Transfer Learning.

With the development of action recognition, applying transfer learning to action recognition datasets generated by different sensors such as visible light cameras, infrared cameras, RGB-D cameras, wearable sensors, or other sensor modalities has received great interests in recent years [33]. As video sequences are the most common type of action datasets, the difference in manipulating and deploying the camera to capture actions leads to different issues, for example, various camera viewing angles, cluttered background, illumination changes, and different light spectrums such as visible light spectrum and infrared spectrum, and all contribute to significant variance in the captured videos. Therefore, action recognition especially cross-domain action recognition is a challenging problem.

Transferring knowledge for cross-view action recognition is prevailing [34]. For example, Zheng et al. [35] proposed learning a pair of dictionaries simultaneously from videos pairs taken at different views to encourage each video pair to have the same sparse representation. Zhang et al. [36] proposed a linear transformation to transform source view to target view via virtual path. Wu et al. [37] proposed a method to discover a discriminative common representation space where source and target views are linked and knowledge is transferred between them. Sui et al. [38] introduced two different projection matrices to map the action data from two different views into the common space with low intraclass diversity and high interclass variance and reducing the mismatch between them. Zu and Zhang [39] introduced a method called Canonical Sparse Cross-view Correlation Analysis to address multiview feature extraction problem. Different from the above-mentioned cross-view action recognition problems, we tackle the visible-to-infrared action recognition problem, which is essentially a cross-dataset action recognition problem.

In recent works about cross-dataset action recognition, Bian et al. [40] proposed a transfer topic model (TTM) which utilized information from the auxiliary domain to assist recognition tasks in the target domain. Zhu and Shao [18] introduced a weakly supervised cross-domain dictionary learning (WSCDDL) approach which learns a reconstructive, discriminative, and domain-adaptive dictionary pair and the corresponding classifier parameters to address cross-domain image classification, action recognition, and event recognition problems. Tang et al. [41] improved the accuracy of action recognition in RGB videos by activating the borrowing of visual knowledge across different video modalities such as RGB videos, the depth maps, and the skeleton data of actions. Liu et al. [42] proposed a simple to complex action transfer learning model (SCA-TLM) to leverage the abundant labeled simple actions to improve the performance of complex action recognition. Although these works can achieve promising results in their related fields, visible-to-infrared action recognition has not attracted much attention until now.

With the revival of neural networks in recent years, many neural networks based transfer learning methods have been proposed as well. Kan et al. [43] proposed bishifting autoencoder network (BAE) to alleviate the discrepancy between source and target domains and evaluate its effectiveness in face recognition. Xu et al. [19] tackled the cross-dataset action recognition problem by training a pair of many-to-one encoders in parallel to map raw features from the source and target datasets to the same space. Although dual many-to-one encoder in [19] can generalize features well across datasets, it requires a large number of labeled training samples from both source and target datasets to learn domain-invariant features without utilizing auxiliary domain data as an aide. In addition, the inputs of the encoders are raw action features named action bank features without considering feature alignment in advance. Different from the above-mentioned neural networks based transfer learning methods, we take both feature alignment and auxiliary domain data into consideration and propose a novel aligned-to-generalized encoder (AGE) model to map the aligned feature representations to the same generalized feature space with low intraclass diversity and high interclass variance.

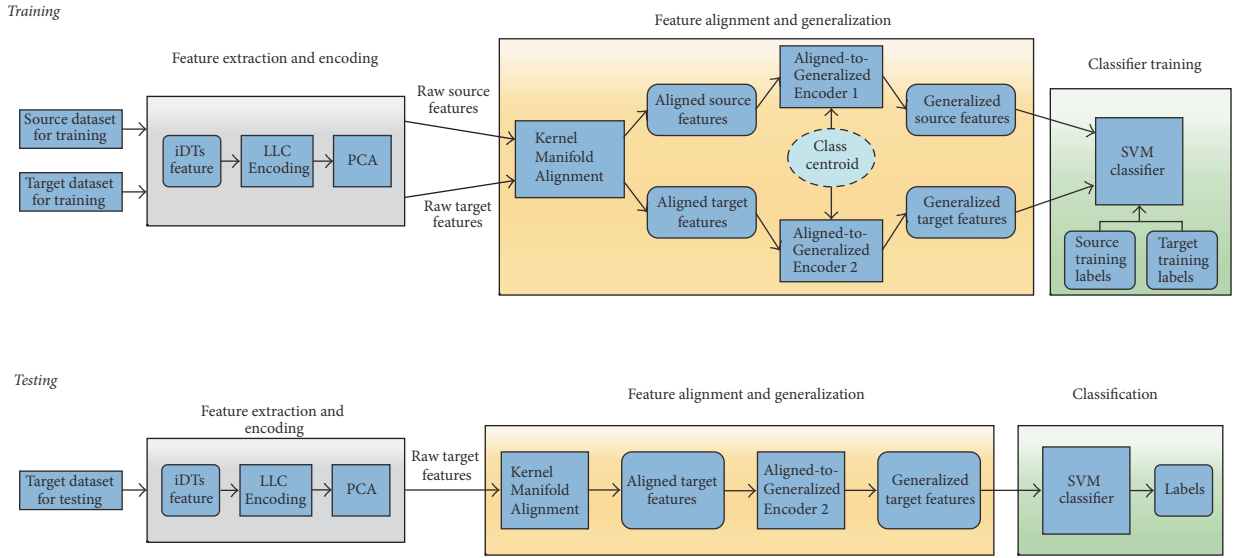

FIGURE 1: The framework of cross-dataset feature alignment and generalization (CDFAG). Top: training. Bottom: testing.

*2.3. Infrared Action Recognition.* Basically, most of the current research efforts for action recognition have been put in visible light videos while infrared action recognition has not attracted much attention. Recently, increasing efforts have been devoted to infrared action recognition. For example, Han and Bhanu [44] proposed an efficient spatiotemporal representation for human repetitive action recognition under thermal infrared scenarios. Han and Bhanu [14] introduced a hierarchical scheme to combine color and thermal images to improve human silhouette detection. Eum et al. [45] used hog and support vector machine to realize infrared action recognition at night. However, these works focus on simple actions under relatively simple environment with limited infrared data. In addition, there is no publicly available infrared action recognition dataset until Gao et al. [13] built the first public infrared action recognition dataset called InfAR. In [13], state-of-the-art action recognition pipelines including widely used low-level local descriptors were evaluated in InfAR dataset. Then Gao et al. [20] extended their previous work [13] and utilized several state-of-the-art pipelines based on low-level features and deep convolutional neural network to evaluate their new infrared action recognition dataset (InfAR). However, the best recognition accuracy in [20] is relatively low and leaves a reasonable space to further promote the performance on InfAR dataset.

Actually, transfer learning has seldom been applied to infrared action recognition. For example, Zhu and Guo [46] proposed applying adaptive support vector machine (A-SVM) [47] to adapt the existing visible light action classifier to classify infrared actions and achieved preliminary results in their own dataset. Although the adaptive support vector machine (A-SVM) based method in [46] can perform better than direct matching, A-SVM is essentially a classifier-transfer based method without considering the max-margin property for the adapted classifier on target instances; therefore it faces accuracy degradation as a result of overfitting.

Our proposed method differs from the above-mentioned approaches in such aspects that it more comprehensively projects and aligns data from different domains in a nonlinear way through kernelization to generate aligned representations and learns a dual aligned-to-generalized encoders to obtain cross-domain generalized features while considering both the discriminability and domain adaptability at the same time. In our proposed CDFAG, the learned classifier across source domain and target domain becomes more discriminative against modality gap because of the integration of both source and target domain knowledge, while a majority of previous transfer learning methods focus on incomplete target domain without utilizing other domain data as an aide to improve the performance of original recognition systems. We will detail our proposed method in Section 3.

## 3. The Proposed Method

In this section, we detail our proposed CDFAG. An overview of the CDFAG is presented in Figure 1. Actually, the proposed CDFAG is semisupervised as both the labeled and the unlabeled data are used in source and target training sets. Our proposed CDFAG consists of three stages. In the first stage, feature extraction and encoding are accomplished on both the source and target datasets, where improved dense trajectories (iDTs) features are extracted, encoded, and reduced to a low-dimensional subspace. In the second stage, aligned features of source and target domains are generated by Kernel Manifold Alignment, then a pair of aligned-to-generalized encoders are trained on the source and target datasets in parallel guided by the centroids of training aligned instances from each class, and after that the output values of the encoders are extracted as the ultimate generalized representations. Finally, a support vector machine is built on the generalized features extracted from both the source and target datasets and then used to classify the new features extracted from unseen samples of the target dataset.

### 3.1. Preprocessing

#### 3.1.1. Feature Extraction and Encoding.
In this paper, we choose improved dense trajectories (iDTs) [48] features with trajectory shape, HOF, MBHx, and MBHy as the low-level action video representation. The total length of the feature vector is 330. Specifically, we use the implementation released on the website of Wang (https://lear.inrialpes.fr/people/wang/improved_trajectories/) for iDTs and choose the default parameter setting. For iDTs, a large number of local trajectory descriptors may lead to high computational complexity and memory consumption. To cope with this issue, we adopt Locality-constrained Linear Coding (LLC) [49] scheme to represent the iDTs by multiple bases, which can bring less quantization error while preserving the local smooth sparsity. Taking both efficiency and the construction error into consideration, LLC encoding scheme is applied to the iDTs with 5 local bases, and the codebook size is set to be 4000 for all training-testing partitions. Thus, the dimension of the encoded iDTs features is 4000. To reduce the complexity when constructing the codebook, only 200 local iDTs are randomly selected from each video sequence.

#### 3.1.2. Principal Component Analysis.
After LLC encoding, the feature representations are still high dimensional and strongly correlated. To obtain more compacted feature representation, we utilize principal component analysis (PCA) [50] to preprocess these features. In our method, we retain top $p$ principal components such that the cumulative corresponding eigenvalues cover over 99% of the total eigenvalues. In our experiments, this reduces feature dimension down to the range of 500 to 600, varying between datasets.

### 3.2. Feature Alignment by Kernel Manifold Alignment.
In this section, we detail the feature alignment method based on Kernel Manifold Alignment (KEMA). An illustration of how feature alignment functions is shown in Figure 2.

#### 3.2.1. Notation.
To fix notation, we consider $K$ input domains. The data instances of each domain belong to $c$ different classes. Let $X_k = (x_k^1, \ldots, x_k^{m_k})$ represent the $k$th input domain, where $m_k$ is the number of samples in the $k$th domain. The idea of kernelization is to map the input data instance into a high dimensional Hilbert space $\mathscr{H}$ with the mapping function $\phi : \mathbf{x} \mapsto \phi(\mathbf{x}) \in \mathscr{H}$ such that the mapped data is better suited for solving our problem linearly. Kernel trick is adopted in our method to avoid high computational load. Therefore, we define a kernel function $\mathbf{K}_{ij} = K(\mathbf{x_i}, \mathbf{x_j}) = \langle \phi(\mathbf{x_i}), \phi(\mathbf{x_j}) \rangle_{\mathscr{H}}$ computing the similarity between mapped instances without having to compute $\phi(\cdot)$ explicitly. Many common types of kernel functions can be adopted in KEMA, such as the RBF kernel, the linear kernel, and the polynomial kernel. In this paper, we use the RBF kernel. Considering multiple data modalities here, we would have to map the $K$ datasets to $K$ Hilbert spaces $\mathscr{H}_k$ of dimension $H_k$, $\phi_k(\cdot) : \mathbf{x} \mapsto \phi_k(\mathbf{x}) \in \mathscr{H}_k$, $k = 1, \ldots, K$.

#### 3.2.2. Kernel Manifold Alignment (KEMA).
The KEMA method aims to construct $K$ domain-specific projection functions, $f_1, f_2, \ldots, f_K$, to project the data in Hilbert space from all $K$ domains to a new common latent space, on which the instances' topology of each domain is preserved, the instances from the same classes will locate nearly, and the ones from different classes will be far from each other. To do so, KEMA aims to find a data projection matrix $F = [f_1, f_2, \ldots, f_K]^\top$ that minimizes the following cost function:

$$\{f_1, f_2, \ldots, f_K\} = \arg \min_{f_1, f_2, \ldots, f_K} (C(f_1, f_2, \ldots, f_K))$$

$$= \arg \min_{f_1, f_2, \ldots, f_K} \left( \frac{\mu \text{TOP} + (1 - \mu) \text{SIM}}{\text{DIS}} \right), \quad (1)$$

where TOP, SIM, and DIS denote the topology, class similarity, and class dissimilarity, respectively. $\mu$ is a parameter balancing the contribution of the similarity and the topology terms. As $\mu \in [0, 1]$, we can see that when $\mu > 0.5$, more importance is given to topology and vice versa. The three terms are defined as follows:

(1) Minimizing a topology-preservation term, TOP, which aims to preserve the local topology of each data domain:

$$\text{TOP} = \sum_{k=1}^{K} \sum_{i,j=1}^{m_k} \left\| \mathbf{f}_k^\top \phi_k(\mathbf{x}_k^i) - \mathbf{f}_k^\top \phi_k(\mathbf{x}_k^j) \right\|^2 W_t^k(i,j)$$

$$= \text{tr}(\mathbf{F}^\top \mathbf{\Phi} \mathbf{L}_t \mathbf{\Phi}^\top \mathbf{F}), \quad (2)$$

where $W_t^k$ in the similarity matrix representing the similarity of $x_k^i$ and $x_k^j$, which can be computed as $e^{-\|x_k^i - x_k^j\|^2}$. $L_t \in \mathscr{R}^{(\Sigma_k m_k) \times (\Sigma_k m_k)}$ is the graph Laplacian matrix issued from $L_t = D_t - W_t$ while $D_t$ is the diagonal row sum matrix defined as $D_t(i,i) = \sum_j W_t(i,j)$.

(2) Minimizing a class similarity term, SIM, which encourages the locations of instances with the same class label to be close with each other in the new latent space:

$$\text{SIM}$$

$$= \sum_{k,k'=1}^{K} \sum_{i,j=1}^{m_k, m_{k'}} \left\| \mathbf{f}_k^\top \phi_k(\mathbf{x}_k^i) - \mathbf{f}_{k'}^\top \phi_{k'}(\mathbf{x}_{k'}^j) \right\|^2 W_s^{k,k'}(i,j) \quad (3)$$

$$= \text{tr}(\mathbf{F}^\top \mathbf{\Phi} \mathbf{L}_s \mathbf{\Phi}^\top \mathbf{F}),$$

where $W_s^{k,k'}$ in the similarity matrix is set to be 1 if two instances from domains $k$ and $k'$ share the same class label and 0 otherwise (including the case when the label information is not available). The corresponding diagonal row sum matrix is defined as $D_s(i,i) = \sum_j W_s(i,j)$ and the graph Laplacian matrix $L_s = D_s - W_s$.

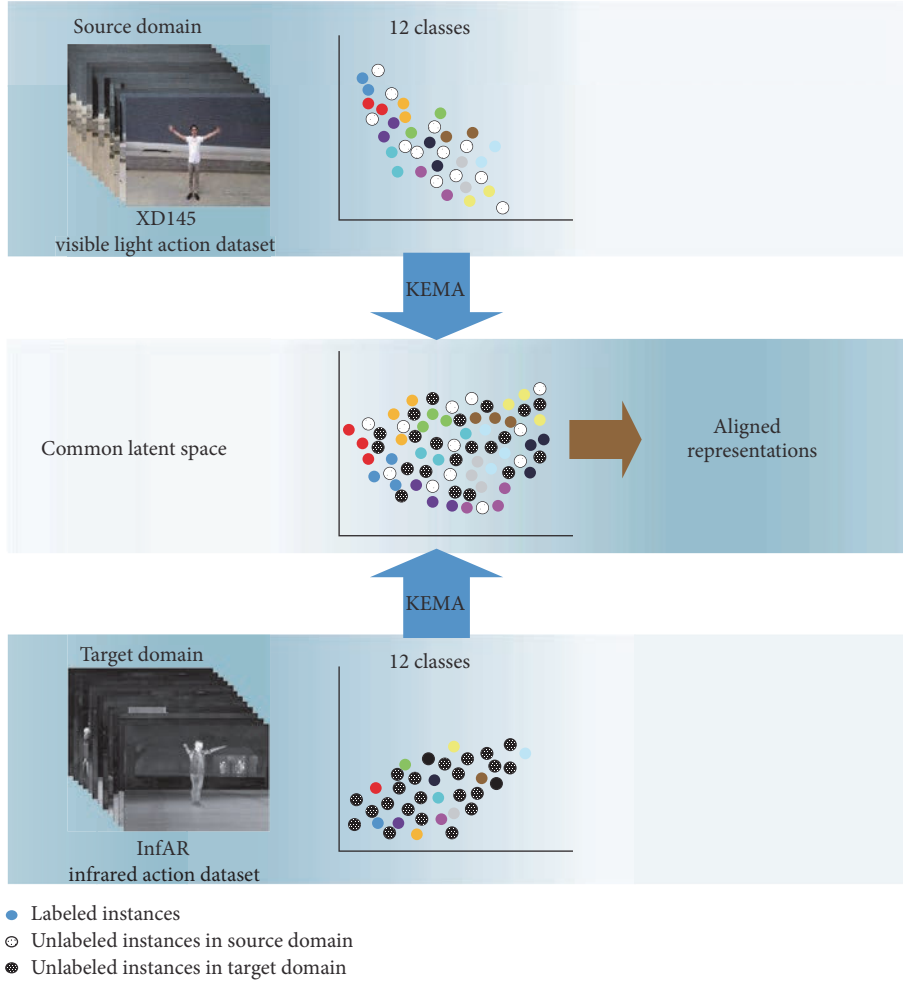

FIGURE 2: Illustration of feature alignment in a visible-to-infrared action recognition setting. Circles in different colors represent different action classes.

(3) Maximizing a class dissimilarity term, DIS, which encourages instances with different class labels to be separated in the new latent space,

DIS

$$= \sum_{k,k'=1}^{K} \sum_{i,j=1}^{m_k, m_{k'}} \left\| \mathbf{f}_k^\top \phi_k\left(\mathbf{x}_k^i\right) - \mathbf{f}_{k'}^\top \phi_{k'}\left(\mathbf{x}_{k'}^j\right) \right\|^2 W_d^{k,k'}\left(i,j\right) \quad (4)$$

$$= \operatorname{tr}\left(\mathbf{F}^\top \boldsymbol{\Phi} \mathbf{L}_d \boldsymbol{\Phi}^\top \mathbf{F}\right),$$

where $W_d^{k,k'}$ in the dissimilarity matrix is set to 1 if two instances from domains $k$ and $k'$ are from different classes and 0 otherwise (including the case when the label information is not available). The corresponding diagonal row sum matrix is defined as

$D_d(i,i) = \sum_j W_d(i,j)$ and the graph Laplacian matrix $L_d = D_d - W_d$.

Given (2)–(4), the optimization problem is formalized as follows:

$$\arg \min_{f_1, f_2, \dots, f_K} \operatorname{tr}\left(\frac{\mathbf{F}^\top \boldsymbol{\Phi}\left(\mu \mathbf{L_t} + (1-\mu)\,\mathbf{L_s}\right) \boldsymbol{\Phi}^\top \mathbf{F}}{\mathbf{F}^\top \boldsymbol{\Phi} \mathbf{L_d} \boldsymbol{\Phi}^\top \mathbf{F}}\right). \quad (5)$$

It is straightforward that the solution of (5) boils down to finding the $n$ lowest eigenvalues of the following generalized eigenvalue decomposition [51]:

$$\boldsymbol{\Phi}\left(\mu \mathbf{L_t} + (1-\mu)\,\mathbf{L_s}\right) \boldsymbol{\Phi}^\top \mathbf{F} = \lambda \boldsymbol{\Phi} \mathbf{L_d} \boldsymbol{\Phi}^\top \mathbf{F}, \quad (6)$$

where $\boldsymbol{\Phi}$ is a matrix containing the matrices $\boldsymbol{\Phi}_k = [\phi_k(\mathbf{x}_k^1), \dots, \phi_k(\mathbf{x}_k^{m_k})]^\top$ in a block diagonal form and $\mathbf{F}$ contains the row eigenvectors for the particular domain defined in Hilbert space $\mathcal{H}_k$, where $\mathbf{F} = [\mathbf{f}_1, \mathbf{f}_2, \dots, \mathbf{f}_H]^\top$, $H = \sum_{k=1}^{K} H_k$, and $\lambda$ is the eigenvalues of the generalized eigenvalue decomposition problem. Note that $\boldsymbol{\Phi}$ and $\mathbf{F}$ are high dimensional and cannot be explicitly computed. Therefore, the eigenvectors are expressed as a linear combination of mapped instances using the Riesz representation theorems [52] and $\mathbf{f}_k = \boldsymbol{\Phi}_k \boldsymbol{\alpha}_k$ and in matrix notation $\mathbf{F} = \boldsymbol{\Phi}\boldsymbol{\Lambda}$. In (6), by multiplying both sides by $\boldsymbol{\Phi}^\top$ and replacing the dot products with the corresponding kernel matrices, $\mathbf{K}_k = \boldsymbol{\Phi}_k^\top \boldsymbol{\Phi}_k$, the final solution is obtained:

$$\mathbf{K}\left(\mu \mathbf{L}_t + (1 - \mu)\,\mathbf{L}_s\right)\mathbf{K}\boldsymbol{\Lambda} = \lambda \mathbf{K}\mathbf{L}_d \mathbf{K}\boldsymbol{\Lambda}, \tag{7}$$

where $\mathbf{K}$ is a matrix containing the kernel matrices $\mathbf{K}_k$ in a block diagonal form. The block structure of projection matrix $\boldsymbol{\Lambda}$ is as follows:

$$\boldsymbol{\Lambda} = \begin{bmatrix} \boldsymbol{\alpha_1} \\ \vdots \\ \boldsymbol{\alpha}_K \end{bmatrix} = \begin{bmatrix} \boldsymbol{\alpha_{1,1}} & \cdots & \boldsymbol{\alpha_{1,n}} \\ \vdots & \ddots & \vdots \\ \boldsymbol{\alpha_{n_1,1}} & \cdots & \boldsymbol{\alpha_{n_1,n}} \\ \alpha_{n_1+1,1} & \cdots & \alpha_{n_1+1,n} \\ \vdots & \ddots & \vdots \\ \alpha_{n,1} & \cdots & \alpha_{n,n} \end{bmatrix}, \tag{8}$$

where the eigenvectors for the first domain are highlighted in bold.

Once the projection matrix $\boldsymbol{\Lambda}$ is obtained, the instance $\mathbf{x}_k^i$ from $k$th domain can be projected to the new latent space by first mapping $\mathbf{x}_k^i$ to its corresponding kernel form $\mathbf{K}_k^i$ and then applying the corresponding projection vector $\boldsymbol{\alpha}_k$ defined therein:

$$P\left(\mathbf{x}_k^i\right) = \mathbf{f}_k^\top \boldsymbol{\Phi}_k^i = \boldsymbol{\alpha}_k^\top \boldsymbol{\Phi}_k^\top \boldsymbol{\Phi}_k^i = \boldsymbol{\alpha}_k^\top \mathbf{K}_k^i, \tag{9}$$

where $\mathbf{K}_k^i$ is a kernel evaluations vector between instance $\mathbf{x}_i$ and all instances from $k$th domain used to define the projections $\boldsymbol{\alpha}_k$. Similar to eigenvalue decomposition based methods, the data can be projected onto a lower-dimensional subspace by simply preserving the first $p$ columns of $\boldsymbol{\alpha}_k$, where $n = \sum_k n_k$ is the total number of samples involved in the kernel matrices and $p \ll n$. In this sense, KEMA leaves some control on the dimensionality of the latent space for feature alignment.

In this paper, the number of input domains $K$ is set to 2 because there is only one target domain (infrared dataset) and one source domain (visible light dataset) in our experiments.

### 3.3. Feature Generalization by Aligned-to-Generalized Encoders.
Due to the huge modality gap between infrared and visible light data, a unified subspace may not exist when only using KEMA to align features from both domains. To be more specific, KEMA holds the assumption that a unified aligned space for both the source and the target domain exists. This assumption is too strict and may be invalid for some cases. Therefore, we relax this strict assumption and learn transferable feature representations across infrared and visible domains in a hierarchical way. With the obtained aligned representation, aligned-to-generalized encoders (AGE) model is adopted to force the outputs to be identical to the input aligned instances from the same action class. The AGE is trained by the guidance of the identical representation of the same action class, where intraclass diversities are minimized and generalized representations are generated across datasets. In this section, we present the architecture and details of the proposed AGE.

### 3.3.1. Target Output Generation.
The centroid of each action class is used as the target output, which is computed by averaging over instances' aligned feature representations in each class. Let $X_{s,c}^i$ and $X_{t,c}^j$ denote the aligned representations of $i$th and $j$th training instances from class $c$ in the source and target dataset, respectively, and the target output for instances from class $c$ is defined as follows:

$$T_c = \frac{1}{N_{s,c} + N_{t,c}}\left(\sum_{i=1}^{N_{s,c}} X_{s,c}^i + \sum_{j=1}^{N_{t,c}} X_{t,c}^j\right), \tag{10}$$

where $N_{s,c}$ and $N_{t,c}$ denote the total number of instances of class $c$ from the source and target datasets.

### 3.3.2. Encoders Training.
At the feature generalization stage, a pair of aligned-to-generalized encoders are trained on the source and target datasets in parallel. For instances, with the same action class label, the target outputs of the two aligned-to-generalized encoders are identical, which forces the two aligned-to-generalized encoders to generalize to varying inputs and guide outputs of the same class instances to be similar. In this sense, aligned instances with the same class label from two datasets are mapped to the same feature space [36, 53] with low intraclass diversity [19] and then generalized and discriminative representations of the instances are generated across datasets.

In this section, we demonstrate the benefits of using a pair of aligned-to-generalized encoders for feature generalization in visible-to-infrared action recognition. The architecture of the aligned-to-generalized encoders (AGE) is illustrated in Figure 3. The AGE are essentially fully connected feedforward neural networks with an input layer, a hidden layer, and an output layer. Although the intuition that a deeper network architecture with more than one hidden layer can learn more robust and discriminative representations, it has been shown that carefully configured and trained single-hidden-layer neural networks can also achieve good performance in many tasks [19, 54], which is validated in our experiments as well. In addition, the architectures of the AGE trained on the source and target datasets are identical.

After feature extraction and encoding, the training videos from both datasets are represented by iDTs features encoded by LLC, which are then reduced to a low-dimensional subspace via PCA. Then, the KEMA method is applied to map the raw feature representations of instances from two datasets to the common latent space to obtain aligned features, shown in Figure 3. For simplicity, we denote one training aligned instance as $x_i$ with dimension $L$. Therefore, both aligned-to-generalized encoders have an input layer of size $L$, a hidden layer of size $H$, and an output layer of size $L$, where $L$ is the length of the ultimate aligned output feature vector and $H$ and $L$ are user defined parameters. In this work, we empirically restrict $H$ to be equal to $L$ and experimented with hidden layer sizes that range from 50 to 500 in Section 4.4.3.

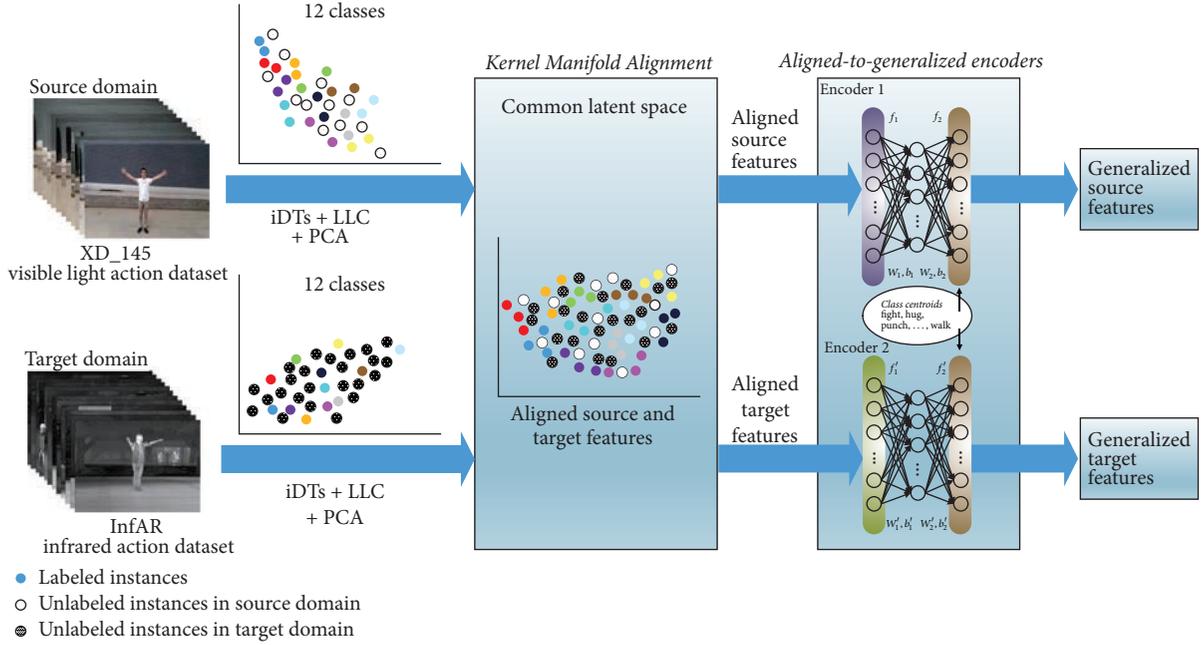

Figure 3: Illustration of aligned-to-generalized encoders generalizing aligned features across visible light and infrared action datasets.

Although both aligned-to-generalized encoders have the same architecture, they have different parameters. The goal of training the two encoders in parallel is to find a mapping between training aligned instances and target generalized outputs. Take the source encoder as an example; the mapping is accomplished via $f_1 : \mathbb{R}^L \to \mathbb{R}^H$ and $f_2 : \mathbb{R}^H \to \mathbb{R}^L$. $f_1$ and $f_2$ are defined as follows:

$$
\begin{aligned}
f_1(x_i) &= s(W_1 x_i + b_1), \\
f_2(f_1(x_i)) &= s(W_2 f_1(x_i) + b_2),
\end{aligned}
\tag{11}
$$

where $s(\cdot)$ is the activation function and $W_1$, $b_1$, $W_2$, and $b_2$ are the parameters for $f_1$ and $f_2$, respectively. The encoder parameters are indicated in Figure 3. Given a hidden layer size $H$, the weights and biases in both encoders are initialized with random numbers drawn from a uniform distribution ranging between 0 and 1. The sigmoid function was used as the activation function:

$$
s(x) = \frac{1}{1 + e^{-x}}.
\tag{12}
$$

The objective functions are defined as follows:

$$
\begin{aligned}
J &= \frac{1}{2N} \sum_{i=1}^{N} \left\| T_i - h_{W_1, b_1, W_2, b_2}(x_i) \right\|_2^2 \\
&= \frac{1}{2N} \sum_{i=1}^{N} \left\| T_i - f_2(f_1(x_i)) \right\|_2^2 \\
&= \frac{1}{2N} \sum_{i=1}^{N} \left\| T_i - s(W_2 s(W_1 x_i + b_1) + b_2) \right\|_2^2,
\end{aligned}
\tag{13}
$$

where $f_1(\cdot)$ and $f_2(\cdot)$ are defined in (11), $W_1$, $W_2$, $b_1$, and $b_2$ are the weights and biases of the encoder, and $N$ is the number of training instances.

Stochastic gradient descent was utilized to minimize the objective function $J$ by iteratively updating the weights and biases. For example, $W_1$ is updated as follows:

$$
\begin{aligned}
\Delta W_1(t+1) &= \eta \frac{\delta J_{t+1}}{\delta W_1} + \rho \Delta W_1(t), \\
W_1(t+1) &= W_1(t) - \Delta W_1(t+1),
\end{aligned}
\tag{14}
$$

where $\Delta W_1(t+1)$ is the update to $W_1$ at the $(t+1)$th iteration, $J_{t+1}$ is the objective function value at the $(t+1)$th iteration, $\eta$ denotes the learning rate, and $\rho$ denotes the momentum. $W_2$, $b_1$, and $b_2$ are updated in a similar way.

In our method, the trained values of the output layer are extracted as the ultimate generalized features. When the objective function $J$ is minimized, the output values of the encoder are approximate solutions staying close to the target outputs instead of being identical to the predefined target outputs. Therefore, the final features extracted from aligned instances of the same action class, from both the source and the target datasets, would lie in the same cluster with small intraclass diversity and high interclass variance. This phenomenon will be illustrated in Section 4.4.2.

To make it clear, the proposed CDFAG is summarized in Algorithm 1.

*3.4. Classification.* Due to the lack of infrared data, directly using neural network for classifying may lead to overfitting. Therefore, we just use the AGE as a feature extractor rather than a classifier to avoid overfitting. In our experiments, we use multiclass support vector machine (SVM) as a classifier

---

**Input:**

Raw features $X = \{X_k \mid X_k = (\mathbf{x}_k^1, \ldots, \mathbf{x}_k^{m_k})\}$, $X_k \in \mathscr{R}^{d_k \times m_k}$, $k = 1, \ldots, K$, the number of input data domains $K$, dimension of common latent subspace $n$, trade-off parameter $\mu = 0.1$, maximum iterations 1000, input layer size $L$, output layer size $L$, hidden layer size $H$, learning rate 0.1, momentum 0.9 and $W_1$, $W_1'$, $b_1$, $b_1'$, $W_2$, $W_2'$, $b_2$, $b_2'$ are randomly initialized.

**Feature alignment:**

(1) Map the raw features from $K$ datasets to $K$ Hilbert spaces:
$$\Phi = \left\{ \Phi_k \mid \Phi_k = \left[ \phi_k\left(\mathbf{x}_k^1\right), \ldots, \phi_k\left(\mathbf{x}_k^{m_k}\right) \right]^\top, \ k = 1, \ldots, K \right\}.$$

(2) Construct graph Laplacian matrices $\mathbf{L}_s$, $\mathbf{L}_s$ and $\mathbf{L}_d$ defined in Section 3.2.2.

(3) Compute the mapping functions $(f_1, f_2, \ldots, f_K)$ by finding the $n$ smallest eigenvalues of the generalized eigenvalue problem:
$$\mathbf{K}\left(\mu \mathbf{L}_s + (1 - \mu)\,\mathbf{L}_s\right)\mathbf{K}\mathbf{A} = \lambda \mathbf{K}\mathbf{L}_d\mathbf{K}\mathbf{A}.$$

(4) Apply $(f_1, f_2, \ldots, f_K)$ to map $K$ input datasets to the new $n$ dimensional common latent space to obtain aligned features:
$$P = \left\{ P\left(\mathbf{x}_k^i\right) \mid P\left(\mathbf{x}_k^i\right) = \mathbf{f}_k^\top \Phi_k^i = \boldsymbol{\alpha}_k^\top \Phi_k^\top \Phi_k^i = \boldsymbol{\alpha}_k^\top \mathbf{K}_k^i, \ k = 1, \ldots, K, \ i = 1, \ldots, m_k \right\}.$$

**Feature generalization:**

(5) Calculate the target outputs of aligned-to-generalized encoders from class $c$:
$$T_c = \frac{1}{N_{s,c} + N_{t,c}} \left( \sum_{i=1}^{N_{s,c}} X_{s,c}^i + \sum_{j=1}^{N_{t,c}} X_{t,c}^j \right),$$

$X_{s,c}^i$ and $X_{t,c}^j$ denote the aligned features of $i$th and $j$th training instances from class $c$ in the source and target dataset.

(6) **for** iter = 1 to 1000 **do**

(7)     Minimize objective function:
$$J = (1/2N) \sum_{i=1}^{N} \|T_i - h_{W_1, b_1, W_2, b_2}(x_i)\|_2^2 \text{ for both encoders in parallel via stochastic gradient descent.}$$

(8) **end for**

(9) Take the activations of aligned-to-generalized encoders as the final generalized features.

**Output:**

Generalized features $X^* = \{X_k^* \mid X_k^* = (x_{k,1}^*, \ldots, x_{k,m_k}^*)\}$, $X_k^* \in \mathscr{R}^{L \times m_k}$, $k = 1, \ldots, K$ across different datasets.

Algorithm 1: Framework of Cross-Dataset Feature Alignment and Generalization (CDFAG).

rather than softmax in AGE because SVM classifier could obtain better results, which has been validated in [55, 56]. To perform visible-to-infrared action classification, a SVM classifier with RBF kernel is trained on generalized features extracted from both visible light (source) and infrared (target) datasets and tested on generalized features extracted from unseen instances from infrared (target) dataset, as shown in the bottom of Figure 1. In Section 4, we will show that such classification scheme can effectively classify unseen action data in target dataset. This can be attributed to the successful knowledge transfer from the source domain to the target domain by our proposed CDFAG. To make it clear, the testing procedure of our proposed CDFAG is summarized in Algorithm 2.

## 4. Experimental Results

In this section, we present our experimental results on the benchmark dataset. We will start with describing the individual datasets, followed by details of our experimental settings.

*4.1. Datasets.* The InfAR dataset (https://sites.google.com/site/gaochenqiang/publication/infrared-action-dataset/) and the XD145 dataset (the dataset will be available at: https://sites.google.com/site/yangliuxdu/) are used for the visible-to-infrared action recognition task, where the XD145 dataset is used as the source domain and the InfAR dataset is used as the target domain.

(A) InfAR

The InfAR dataset [20] consists of 600 video sequences captured by infrared thermal imaging cameras. As shown in Figure 4, fight, handclapping, handshake, hug, jog, jump, punch, push, skip, walk, wave 1 (one-hand wave), and wave 2 (two-hand wave) are included in the dataset, where each action class has 50 video clips and each clip lasts 4 s in average. The frame rate is 25 fps and the resolution is $293 \times 256$. Each video contains one or multiple actions performed by one or several persons. Some of them are interactions between multiple persons, shown in Figure 4.

(B) XD145

We build a visible light action dataset, named XD145, following the approach to construct an action recognition dataset from the visible spectrum [57]. In correspondence with the target domain action categories, both the XD145 and the InfAR dataset have the same action categories, as shown in Figure 5. The XD145 action dataset consists of 600 video sequences captured by visible light cameras and there are 50 video clips for each action class. All actions were performed by 30 different volunteers. Each clip lasts for 5 s in average. The frame rate is 25 fps and the resolution is $320 \times 240$. As shown in Figure 5, the background, pose, and viewpoint variations are considered when constructing the dataset in order to make our dataset more representative for real-world scenarios.

Figure 6 illustrates sample actions from the InfAR and the XD145 datasets. As can be seen in the figure, these action videos are captured in two different light spectra and they

**Input:**

   Raw features of testing samples in target dataset $X_{\text{target,test}} = \{\mathbf{x}_1, \ldots, \mathbf{x}_{N_{\text{test}}}\}$, $X_{\text{target,test}} \in \mathcal{R}^{d \times N_{\text{test}}}$, the number of testing samples
   in target dataset $N_{\text{test}}$, dimension of raw features $d$, dimension of common latent subspace $n$, trade-off parameter $\mu = 0.1$,
   learned projection function for target dataset $\mathbf{f}_{\text{target}}$, trained aligned-to-generalized encoder for target dataset
   $f_1^{\text{target}}$ and $f_2^{\text{target}}$, SVM classifier trained by samples from both source and target dataset.

   **Feature alignment:**

(1) Map the raw features in target datasets to the Hilbert space:

   $$\boldsymbol{\Phi}_{\text{target,test}} = \left[\phi\left(\mathbf{x}_1\right), \ldots, \phi\left(\mathbf{x}_{N_{\text{test}}}\right)\right]^{\top}.$$

(2) Apply $\mathbf{f}_{\text{target}}$ to map the raw features in target datasets to the new $n$ dimensional common latent space to obtain aligned features:

   $$\mathbf{X}_{\text{target,aligned}} = \mathbf{f}_{\text{target}}^{\top} \boldsymbol{\Phi}^i = \boldsymbol{\alpha}^{\top} \boldsymbol{\Phi}_{\text{target,test}}^{\top} \boldsymbol{\Phi}_{\text{target,test}}^i = \boldsymbol{\alpha}^{\top} \mathbf{K}_{\text{target,test}}^i, \; i = 1, \ldots, N_{\text{test}}.$$

   **Feature generalization:**

(3) Input the aligned features $\mathbf{X}_{\text{target,aligned}}$ into the trained aligned-to-generalized encoder and obtained the generalized features
   $\mathbf{X}_{\text{target,generalized}}$ at the output layer:

   $$\mathbf{X}_{\text{target,generalized}} = f_2^{\text{target}}\left(f_1^{\text{target}}\left(\mathbf{X}_{\text{target,aligned}}\right)\right)$$

   **Classification:**

(4) Adopt the trained SVM classifier to predicts the class labels of testing samples in target dataset using generalized features
   $\mathbf{X}_{\text{target,generalized}}$.

**Output:**

   Predicted labels of testing samples in the target dataset.

ALGORITHM 2: Testing algorithm of CDFAG.

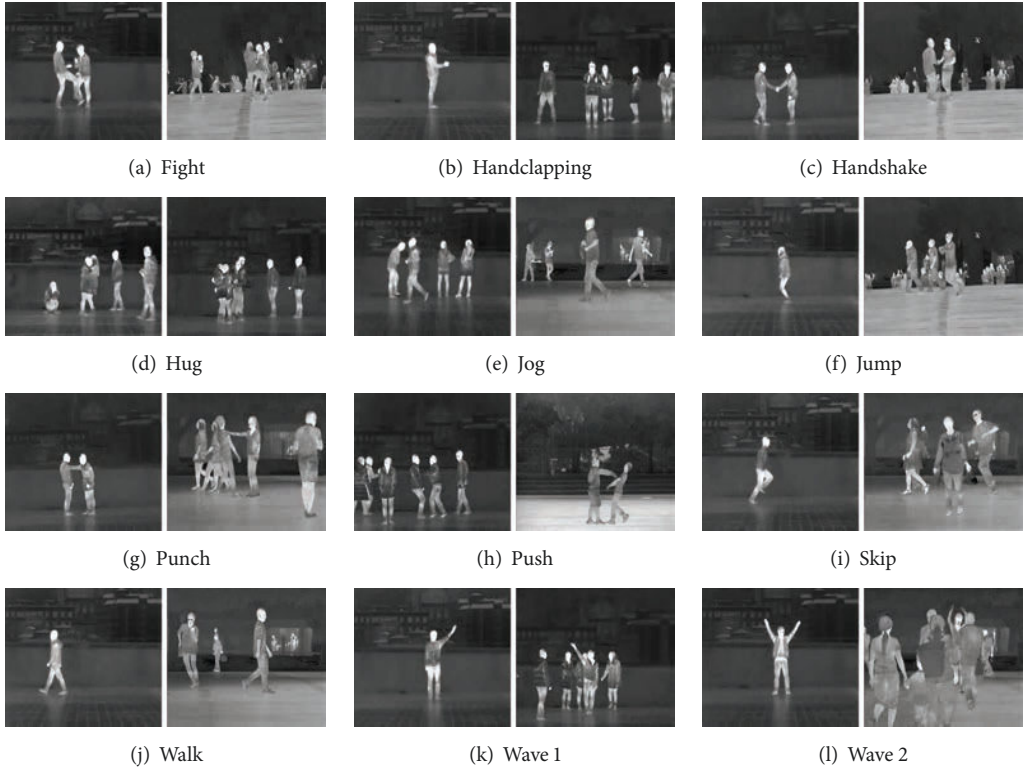

(a) Fight                     (b) Handclapping               (c) Handshake

(d) Hug                       (e) Jog                        (f) Jump

(g) Punch                     (h) Push                       (i) Skip

(j) Walk                      (k) Wave 1                     (l) Wave 2

FIGURE 4: Example images from video sequences in the InfAR dataset.

exhibit significantly great intraclass variance and modality gap.

### 4.2. Experimental Settings.

In all experiments, each dataset is randomly split into training and testing sets. For evaluation, the average precision (AP) is used, which is the average of recognition precisions of all actions. For each evaluation, we repeat the experiments with the same setting 5 times and report the average accuracy. In KEMA, we use the RBF kernels with the bandwith fixed as half of the median distance between the samples of the specific video (labeled and unlabeled). By doing so, we allow different kernels in each domain, thus tailoring the similarity function to the data structure observed [15]. The trade-off parameter $\mu$ is set as 0.1 according to the result of experimental analysis in Section 4.4.3. To build the graph Laplacians, we use a $k$-NN graph with $k = 10$. We validate the optimal dimension $n$ of common latent space as well as the optimal $C$ and $\gamma$ parameters in the SVM classifier. Since the RBF kernel is adopted in KEMA, the LibSVM [58] is used to train classifiers in our experiments with RBF kernel as well. The optimal parameters $C$ and $\gamma$ are determined by 5-fold cross-validation. When performing stochastic gradient descent for encoder training, we set learning rate $\eta = 0.1$ and momentum $\mu = 0.9$ (see (14)). Each encoder is trained for about 1000 iterations. All experiments are conducted with MATLAB R2016b on a 64-bit Windows 10 PC with 4-core 3.60 GHz Intel i7 CPU and 16 GB of memory.

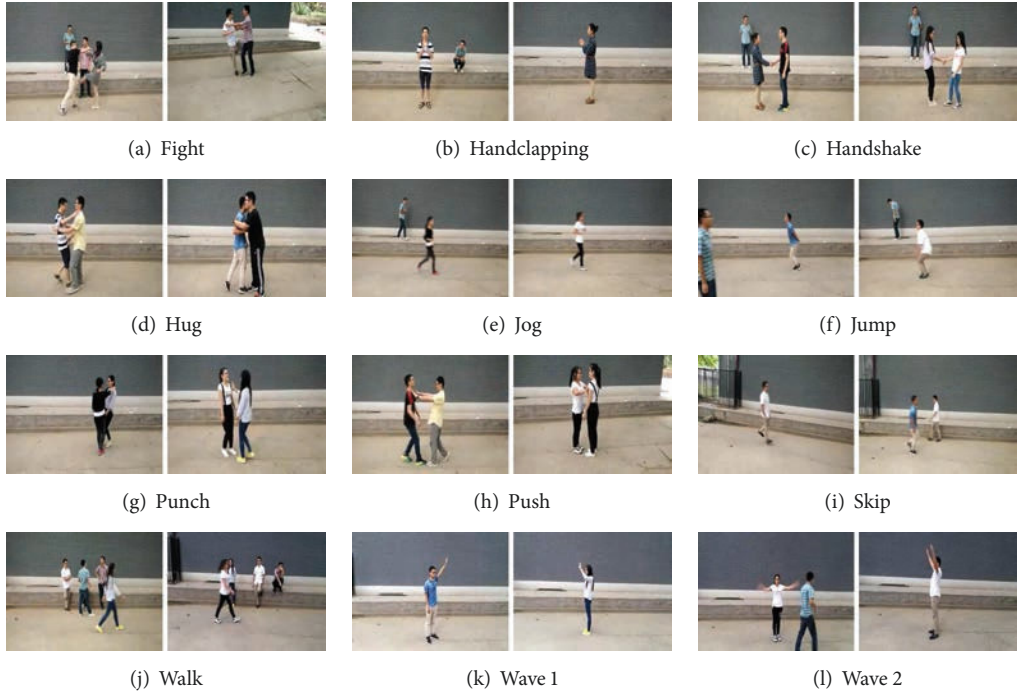

(a) Fight      (b) Handclapping      (c) Handshake

(d) Hug      (e) Jog      (f) Jump

(g) Punch      (h) Push      (i) Skip

(j) Walk      (k) Wave 1      (l) Wave 2

Figure 5: 12 actions of the newly constructed visible light action dataset XD145.

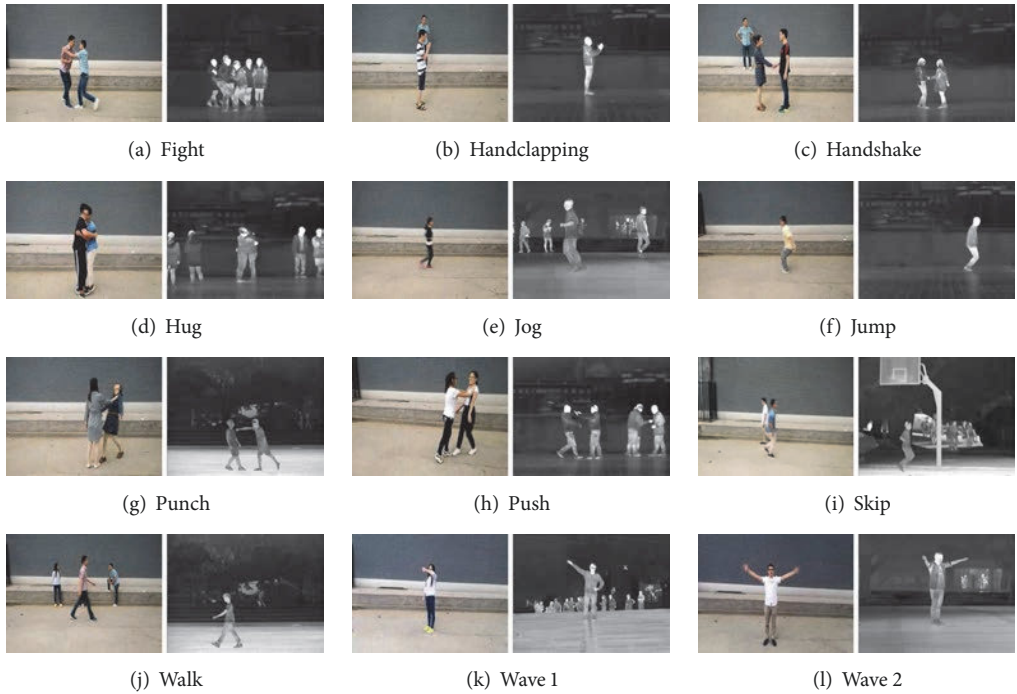

(a) Fight      (b) Handclapping      (c) Handshake

(d) Hug      (e) Jog      (f) Jump

(g) Punch      (h) Push      (i) Skip

(j) Walk      (k) Wave 1      (l) Wave 2

Figure 6: Comparison of visible and infrared actions. The left of each subfigure is the visible image from video sequences in the XD145 dataset, while the right is the infrared image from video sequences in the InfAR dataset.

Table 1: The average recognition precision (%) of iDTs feature in XD145 and InfAR datasets. train_num and test_num denote the number of samples in training and testing sets, respectively.

| Settings | Accuracies in XD145 (%) | Accuracies in InfAR (%) |
|---|---|---|
| train_num = 25, test_num = 20 | 90.58 ± 2.14 | 69.25 ± 2.71 |
| train_num = 20, test_num = 20 | 90.00 ± 2.19 | 66.92 ± 1.37 |
| train_num = 15, test_num = 20 | 86.83 ± 1.78 | 63.92 ± 2.24 |
| train_num = 10, test_num = 20 | 83.58 ± 2.74 | 40.33 ± 7.88 |
| train_num = 5, test_num = 20 | 77.25 ± 5.29 | 30.25 ± 5.41 |

*4.3. Action Recognition Results with Raw Features.* We evaluate the original feature in our newly constructed visible light action dataset XD145 and infrared action recognition dataset InfAR, respectively, which is constructed as baseline in this paper. For each evaluation, we repeat the experiments with the same settings 5 times, where, for each time, the numbers of 25/20/15/10/5 samples out of 50 samples for each action category are randomly selected as the training set and 20 samples out of the remaining samples are used as the testing set. Then, the averages are employed as the final result, as shown in Table 1.

From Table 1, we can observe that the recognition accuracies in both datasets are growing with the number of training samples increasing. This is basically in accordance with the traditional action recognition intuition that a good action recognition result can be achieved with adequate amount of labeled training samples and discriminative features. Although the same feature is adopted in both datasets, the accuracies in InfAR are much lower than that in XD145 even with the same experimental setting. This result may be due to the fact that the videos in these two datasets are captured by different sensors and they exhibit variant appearance and motion information for the same action class, as shown in Figure 6. Since the iDTs feature is good at appearance and motion description in visible light action videos [48], its effectiveness and strength may not be revealed in infrared videos. Therefore, utilizing existing visible light action data as an aide for enhancing infrared action recognition system is urgently needed.

*4.4. Visible-to-Infrared Action Recognition Results.* In this section, we evaluate our proposed method on visible-to-infrared action recognition. Our experiments are mainly divided into four parts. Firstly, a classifier trained on instances from both source and target datasets without feature alignment and generalization is utilized to predict actions in target dataset. Secondly, the classification results with aligned features obtained by KEMA are also provided. Then, we evaluate our proposed CDFAG in visible-to-infrared action recognition. Lastly, we compare our proposed CDFAG with several state-of-the-art methods.

*4.4.1. Visible-to-Infrared Action Recognition.* To find out whether modality gap is existent between the source and target datasets, we first train a classifier using the samples from both source and target datasets without feature alignment and generalization to predict actions in target dataset. We call this method No Adaptation (NA). In addition, the classification results with aligned features obtained by KEMA are provided to show the effectiveness of feature alignment. For each evaluation we repeat the experiments with the same parameter settings 5 times, where, for each time, the numbers of 45/40/30/25/20/15/10/5 samples for each action category in source dataset combined with the number of 25/20/15/10/5 samples for each action category in target dataset are randomly selected as the training set, in order to validate the impact of the number of training samples on recognition accuracy. Then, 20 samples for each action category in target dataset are used as the testing set. The averages are employed as the final result, as shown in Table 2.

We can see that, in Table 2, the best accuracy of the NA in each column (marked bold) occurs when the number of training samples in source dataset (S_train) is relatively small. Comparing the results of infrared action recognition in Table 1 and the NA in Table 2, we can discover that the No Adaptation (NA) method results in better performance than the baseline method only when fewer samples are used for training (T_train = 10 and T_train = 5), which demonstrates that directly transferring the knowledge from the source domain to target domain without considering their divergence can cause significant performance degeneration especially when the number of source domain samples is large. When the number of source and target training samples is small, the performance gets slightly better because of the complementary information between the source and target training samples. However, with the number of source and target training samples increasing, the modality gap between datasets begins dominating the recognition accuracy; then the performance degeneration becomes more serious. Therefore, it is necessary to reduce the modality gap before directly using them together.

Then, we evaluate our proposed CDFAG in visible-to-infrared action recognition. At feature alignment stage, we use the labeled and unlabeled samples to extract KEMA projections and then project all videos in the latent space to obtain aligned samples. We set the dimension of features in common latent space to be 100. We experiment with various feature dimensions and report the results in Section 4.4.3. For videos from the same action class in source dataset XD145, we randomly choose the number of 45/40/35/30/25/20/15/10/5 videos as the training set while using the number of 5 videos as the unlabeled set. For videos from the same action class in target dataset InfAR, 20 samples out of 50 samples are randomly selected as the testing set; then we randomly choose the number of 25/20/15/10/5 videos as the training set while using the remaining videos as the unlabeled set, since the target dataset usually has more unlabeled samples but less labeled samples than the source dataset in real-world scenarios. The unlabeled samples are utilized to compute the graph Laplacians. At feature generalization stage, all the aligned samples from both of the source and target training sets are used to guide the training of the two aligned-to-generalized encoders. After feature generalization, a SVM classifier with RBF kernel is trained on all the aligned and generalized labeled training samples from both source and target datasets. The trained SVM is used to predict all the testing videos in target dataset. All evaluation results are listed in Table 2.

Table 2: The average recognition precision (%) of No Adaptation (NA), KEMA, and the proposed CDFAG in target dataset InfAR. $S\_train$ and $T\_train$ denote the number of training samples in source and target datasets, respectively. Best accuracies of each column for each method are marked in italics.

| Settings | $T\_train = 25$ | $T\_train = 20$ | $T\_train = 15$ | $T\_train = 10$ | $T\_train = 5$ |
|---|---|---|---|---|---|
| $S\_train = 45$ | | | | | |
| NA | 59.67 ± 2.22 | 56.58 ± 1.99 | 52.58 ± 1.99 | 41.33 ± 2.56 | 32.08 ± 2.59 |
| KEMA | 66.08 ± 2.56 | 64.00 ± 3.57 | 60.25 ± 2.74 | 56.08 ± 3.89 | 43.08 ± 4.98 |
| CDFAG | 72.50 ± 1.98 | 69.25 ± 3.71 | 67.66 ± 3.78 | 62.25 ± 2.84 | 47.67 ± 2.91 |
| $S\_train = 40$ | | | | | |
| NA | 63.25 ± 1.85 | 54.00 ± 2.68 | 50.58 ± 2.68 | 42.75 ± 3.57 | 31.25 ± 3.99 |
| KEMA | *69.67 ± 1.43* | 66.25 ± 2.52 | 59.50 ± 2.56 | 53.58 ± 2.58 | 44.50 ± 2.07 |
| CDFAG | *75.42 ± 3.13* | 71.33 ± 1.99 | 67.58 ± 1.73 | *62.42 ± 2.59* | 49.33 ± 3.17 |
| $S\_train = 35$ | | | | | |
| NA | 60.67 ± 2.48 | 57.42 ± 4.05 | 50.00 ± 2.86 | 45.75 ± 2.23 | 32.42 ± 4.79 |
| KEMA | 66.67 ± 1.86 | 64.00 ± 3.92 | *62.67 ± 2.76* | *56.92 ± 2.74* | 41.58 ± 4.14 |
| CDFAG | 73.25 ± 1.43 | 69.25 ± 2.19 | 65.83 ± 2.37 | 61.33 ± 2.51 | 50.06 ± 4.60 |
| $S\_train = 30$ | | | | | |
| NA | 62.58 ± 3.10 | 58.00 ± 2.49 | 52.17 ± 2.75 | 44.42 ± 2.27 | 33.17 ± 1.52 |
| KEMA | 68.75 ± 1.86 | 65.42 ± 2.30 | 57.67 ± 2.20 | 55.58 ± 4.16 | 45.75 ± 4.54 |
| CDFAG | 73.83 ± 2.19 | 69.25 ± 2.35 | 67.25 ± 2.74 | 60.58 ± 2.77 | 50.42 ± 3.90 |
| $S\_train = 25$ | | | | | |
| NA | 61.67 ± 4.54 | 55.92 ± 3.52 | 49.33 ± 2.51 | 46.08 ± 2.87 | 32.33 ± 2.82 |
| KEMA | 67.75 ± 2.08 | 64.17 ± 3.00 | 60.08 ± 1.78 | 52.50 ± 2.28 | 44.92 ± 1.94 |
| CDFAG | 72.25 ± 2.18 | *71.58 ± 2.42* | *68.08 ± 3.87* | 60.92 ± 2.15 | *50.75 ± 2.19* |
| $S\_train = 20$ | | | | | |
| NA | 64.17 ± 1.14 | 59.50 ± 3.08 | 52.42 ± 3.25 | 46.25 ± 2.80 | 33.25 ± 4.42 |
| KEMA | 69.67 ± 2.80 | 67.00 ± 3.15 | 62.58 ± 2.63 | 52.42 ± 3.43 | *46.25 ± 2.38* |
| CDFAG | 71.50 ± 2.22 | 70.58 ± 2.61 | 66.92 ± 4.10 | 60.92 ± 3.72 | 47.00 ± 6.08 |
| $S\_train = 15$ | | | | | |
| NA | 63.50 ± 3.21 | 59.75 ± 5.28 | *54.17 ± 4.31* | 46.42 ± 2.35 | *33.83 ± 3.22* |
| KEMA | 67.75 ± 3.90 | *67.08 ± 3.27* | 56.33 ± 2.45 | 55.33 ± 1.54 | 42.25 ± 4.13 |
| CDFAG | 70.75 ± 2.09 | 69.16 ± 1.79 | 66.00 ± 2.65 | 59.58 ± 2.86 | 47.92 ± 4.59 |
| $S\_train = 10$ | | | | | |
| NA | *66.50 ± 2.01* | *59.75 ± 3.14* | 54.00 ± 3.63 | *46.75 ± 3.53* | 33.58 ± 4.11 |
| KEMA | 69.17 ± 3.55 | 66.83 ± 3.64 | 62.58 ± 2.64 | 54.00 ± 5.42 | 38.83 ± 2.97 |
| CDFAG | 71.67 ± 2.43 | 67.42 ± 1.54 | 64.92 ± 2.09 | 61.08 ± 3.14 | 45.92 ± 3.90 |
| $S\_train = 5$ | | | | | |
| NA | 66.33 ± 1.87 | 53.42 ± 11.15 | 45.92 ± 11.56 | 33.92 ± 2.87 | 30.67 ± 6.82 |
| KEMA | 65.00 ± 2.04 | 65.75 ± 4.20 | 60.17 ± 3.53 | 56.17 ± 3,34 | 35.67 ± 4.25 |
| CDFAG | 72.00 ± 2.72 | 68.58 ± 3.03 | 65.67 ± 2.77 | 57.00 ± 2.85 | 45.08 ± 3.08 |

Comparing the results of the NA and the KEMA in Table 2, it is obvious that the KEMA achieves higher accuracies than that of the NA under all parameter settings which validates the effectiveness of KEMA in aligning the features across the source and target domains. In addition, it is obvious that the infrared action recognition accuracies of our proposed CDFAG are significantly much higher than that of No Adaptation (NA) and the KEMA under all parameter settings, which validates the effectiveness of our proposed CDFAG in reducing the modality gap between the source and target datasets by using both feature alignment and generalization. Furthermore, we achieve at least 5% increase in infrared action recognition accuracy over the baseline method in Table 1 (the third column) as a result of utilizing source domain data as an aide for enhancing the present recognition system by a novel feature alignment and generalization method. In order to have a more intuitive comparison, best accuracies under different number of target training samples for each method are plotted in Figure 7. As can be illustrated in Figure 7, our proposed method achieves remarkable performance improvement in infrared action recognition under all parameter settings especially with fewer training samples, which verifies its effectiveness in utilizing auxiliary source domain data under scenario of scarce target training data.

Table 3: The average recognition precision (%) in InfAR with different trade-off parameter $\mu$.

| | $\mu = 0$ | $\mu = 0.1$ | $\mu = 0.2$ | $\mu = 0.3$ | $\mu = 0.4$ | $\mu = 0.5$ | $\mu = 0.6$ | $\mu = 0.7$ | $\mu = 0.8$ | $\mu = 0.9$ | $\mu = 1$ |
|---|---|---|---|---|---|---|---|---|---|---|---|
| $S\_train = 45$; $T\_train = 25$ | 46.39 | 72.75 | 71.83 | 71.17 | 71.17 | 69.50 | 70.58 | 67.08 | 66.33 | 61.58 | 61.08 |
| $S\_train = 40$; $T\_train = 25$ | 45.83 | 73.83 | 70.42 | 72.42 | 72.08 | 68.92 | 67.17 | 66.83 | 66.42 | 64.83 | 56.67 |
| $S\_train = 35$; $T\_train = 25$ | 43.33 | 72.75 | 71.42 | 72.08 | 70.08 | 70.17 | 71.25 | 70.92 | 65.17 | 63.08 | 53.75 |

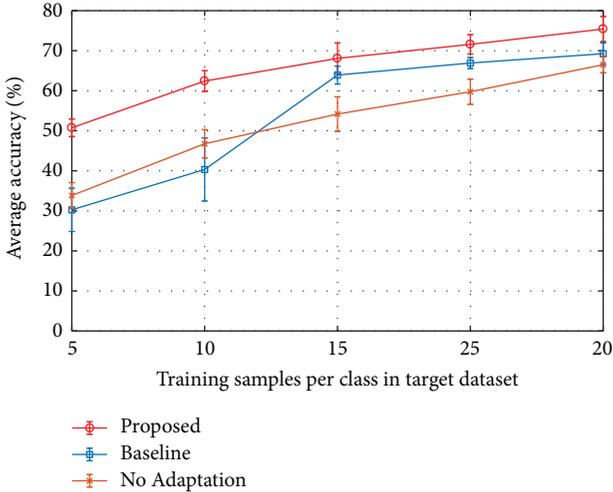

Figure 7: Performance comparison of our proposed method with the baseline method and the No Adaptation (NA) method on the InfAR dataset.

To further explore the infrared action classification performance, two confusion matrices are illustrated in Figure 8. It can be seen that our proposed method achieves higher classification accuracies for nearly all action categories compared with the baseline method. However, there are limited accuracies improvement in two actions—push and punch—since these two actions are easily confused with each other. From Figures 4(g) and 4(h), we can see that punch and push are so similar that it may even be difficult for a human to distinguish them from each other.

### 4.4.2. Visualization of Aligned and Generalized Features.

To verify that, by using our proposed CDFAG, all action data from different datasets are indeed projected into the unique common latent space, we plotted the distribution of the raw features, aligned features, and generalized features of instances from all action classes in source and target datasets, respectively. For illustration purposes, we compared the first 3-dimensional feature space of the raw iDTs features, aligned features, and the learnt generalized features, shown in Figure 9. As can be seen in Figure 9(a), the original features of instances from two datasets are obviously in different clusters with large intraclass diversity and small interclass variance, but they are projected into a single cluster where the instances from the same class are projected to similar locations and the instances from different classes are well-separated, as illustrated in Figure 9(b). Although there is small distinction between Figures 9(b) and 9(c) except that the instances from the same class in Figure 9(c) merge into a more compact cluster, feature generalization indeed maps instances from different datasets to a more compact feature space with relatively small intraclass diversity and large interclass variance, which makes features more generalized and discriminative for visible-to-infrared action recognition.

### 4.4.3. Analysis of Trade-Off Parameter $\mu$.

To evaluate the optimum value range of trade-off parameter $\mu$ in (1), we evaluate the performance of our proposed CDFAG with different values of $\mu$. Specifically, we use $\mu = \{0, 0.1, 0.2, 0.3, 0.4, 0.5, 0.6, 0.7, 0.8, 0.9, 1\}$ in InfAR dataset when (1) $S\_train = 45$ and $T\_train = 25$, (2) $S\_train = 40$ and $T\_train = 25$, and (3) $S\_train = 35$ and $T\_train = 25$. The results of other $S\_train$ and $T\_train$ settings are not evaluated because other settings show similar results to these three settings. For each evaluation, we repeat the experiments with the same settings 5 times and report the average accuracies.

The experimental results are given in Table 3. As shown in Table 3, the average accuracies of these three settings are better when trade-off parameter $\mu$ is small, which indicates that a good performance can be achieved when we attach more importance to class similarity minimization term than topology preservation in KEMA procedure. When we treat both terms equally ($\mu = 0.5$), the performance is also unsatisfactory. These attribute to the fact that the modality difference between infrared and visible light domain is so huge that the topology between them also varies a lot. Therefore, more importance should be attached to class similarity term to achieve good performance. However, when $\mu$ is set as 0, the overall performance drops dramatically. This shows that the topology preservation plays an important role in KEMA and cannot be neglected. When $\mu$ is set as 1, the performance also drops but with less accuracy gap than the case when $\mu = 0$. This means that topology preservation contributes more to the overall performance than class similarity minimization although both of them are necessary in KEMA.

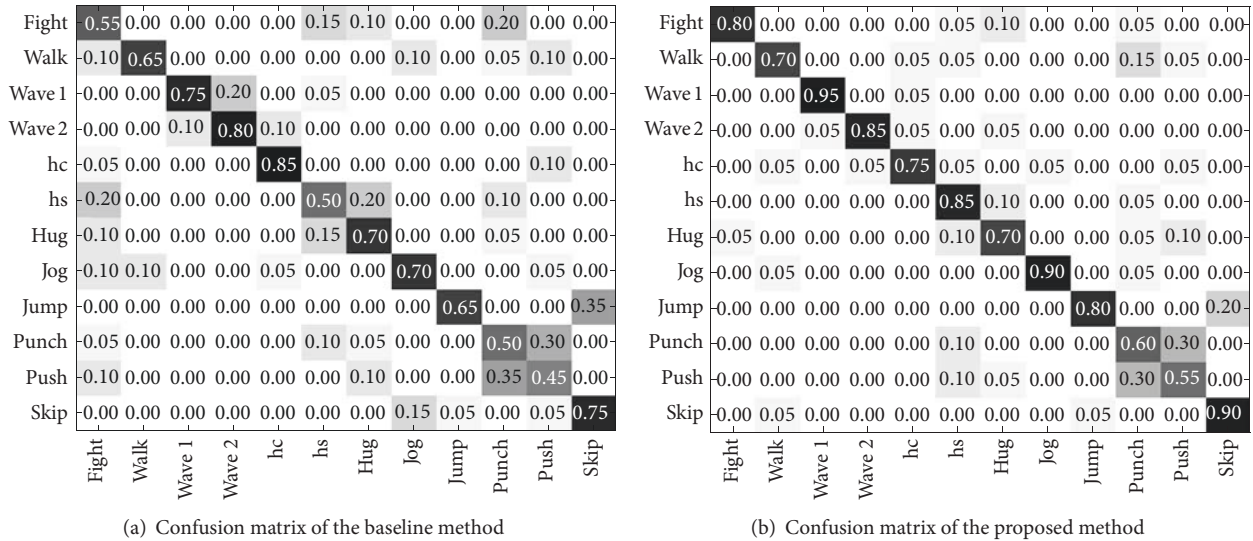

(a) Confusion matrix of the baseline method

(b) Confusion matrix of the proposed method

FIGURE 8: The comparison of the confusion matrices between the baseline method and the proposed method in InfAR dataset, where the left is from the baseline method, while the right is from our proposed CDFAG. Note that "hc" denotes "handclapping" and "hs" denotes "handshaking," respectively.

More intuitive analysis is plotted in Figure 10. From Figure 10 we can see that a good result can be achieved when $\mu < 0.5$ and the optimum value of $\mu$ is 0.1. Therefore, we set $\mu = 0.1$ in the experiments. In addition, the performance is relatively stable with small accuracy gap when $\mu < 0.5$, which shows that our algorithm is insensitive to the trade-off parameter $\mu$ when $\mu < 0.5$.

### 4.4.4. Feature Dimension in Common Latent Space.
We experiment with various feature dimensions $n$ in common latent space to study how the feature dimension influences the classification accuracy. As the hidden layer size $H$ in aligned-to-generalized encoders is empirically set to be equal to the input layer size $L$, the hidden layer size $H$ is directly determined by the value of the feature dimension $n$ in latent space. As illustrated in Figure 11, our proposed method arrives at its best accuracy when feature dimension $n = 100$; then classification accuracy tends to decrease as the feature dimension increases. This can be explained by the decreased discriminability of the feature representations in common latent space as its dimension increases.

### 4.4.5. Computation Time.
We evaluate the computation time of our proposed method and report the results in Table 4. All experiments are conducted on our lab PC and developed by MATLAB. The reported times are averaging running time over all the $S\_train$ values when the $T\_train$ remains unchanged, and the computation time for iDTs features and LLC encoding is not included. The reported time includes feature alignment, feature generalization, classifier training, cross-validation and classification. As can be seen from Table 4, our proposed method can perform feature alignment, feature generalization, classifier training, and classification very efficiently. The longest computation time is just nearly 9 minutes when $n = 500$ and $T\_train = 25$. We attribute this efficient execution to the PCA dimension reduction of raw features, fast feature alignment by introducing the kernel trick in KEMA, and the shallow single-hidden-layer neural network architecture for feature generalization.

### 4.4.6. Comparison with State-of-the-Art Methods.
We compare our proposed CDFAG method with the following state-of-the-art methods:

(i) Domain adaptation based methods: KEMA [15], SSMA [16], and DA [17].

(ii) Transfer learning based methods: WSCDDL [18] and Dual [19].

(iii) Deep learning based methods: two-stream CNNs [20].

Actually, we focus on the transferable features generation in this paper and only use the features generated by these comparison methods including KEMA, SSMA, DA, WSCDDL, and Dual. And then we use SVM as the classifier to recognize the actions. In order to compare with these domain adaptation and transfer learning methods, we use the same experimental setting as our proposed CDFAG (Section 4.4.1) in these related works. Results are reported in Table 5. The results in boldface in each column shows that the proposed CDFAG is the most competitive one compared with other state-of-the-art methods. For instance, the average accuracy of the proposed method CDFAG brings about 20.50%, 22.09%, 4.16%, 4.66%, and 6.17% improvements over the

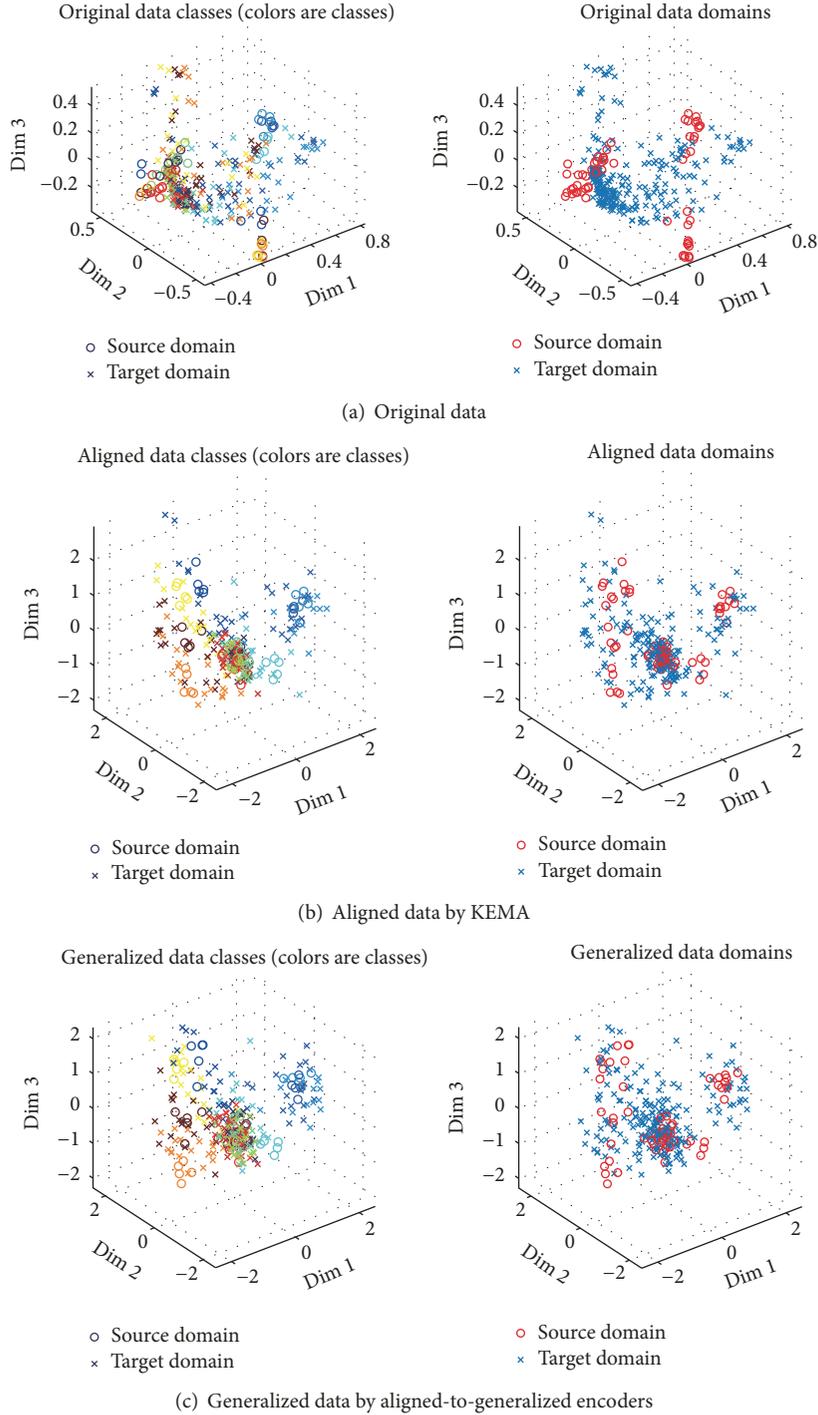

(a) Original data

(b) Aligned data by KEMA

(c) Generalized data by aligned-to-generalized encoders

FIGURE 9: Examples of the three first dimensions of the original, aligned, and generalized data space. The left panel shows data distributions of two domains per class while the right panel shows the data distributions of all classes per domain.

baseline method for five different $T\_$train values, respectively. This validates the effectiveness of the CDFAG in improving the overall infrared action recognition accuracy with the aid of the visible light data from source domain. Compared with domain adaptation methods, the average accuracy of the CDFAG is about 6.00%, 6.59%, and 11.75% higher than that of KEMA [15], SSMA [16], and DA [17] when $T\_$train = 25, respectively. For transfer learning methods, our proposed method also brings about 8.00% and 17.17% improvements over WSCDDL [18] and Dual [19] when $T\_$train = 25, respectively. More comparisons are plotted in Figure 12.

The results in Figure 11 show that the proposed CDFAG performs much better than other state-of-the-art methods on visible-to-infrared action recognition task. As can be seen from Figure 11, the performance of Dual [19] is not good especially with fewer target training samples because it needs adequate training samples to be well-trained. On the one hand, other state-of-the-art methods can achieve higher accuracies than that of the baseline method with fewer target training samples, while, with more target training samples, modality gap begins dominating the recognition accuracies and their performances become inferior to that of the baseline. On the other hand, our proposed method CDFAG can perform well whenever the number of training samples is small or large, which validates the effectiveness of our proposed method CDFAG in terms of reducing the modality gap across the source and target datasets. The key difference between our proposed CDFAG and the other methods is that our method takes both feature alignment and feature generalization into consideration. Thus, a latent common feature space with low intraclass diversity and high interclass variance can be learnt. The other methods focus on only either of them (feature alignment or generalization) without explicit knowledge transfer and effective modality gap reduction. The strong performance of our proposed CDFAG demonstrates the advantage of transferring knowledge from visible light domain to infrared domain and, more importantly, the efficacy of the proposed cross-dataset feature alignment and generalization (CDFAG) framework.

Table 4: Average computation time of our proposed method CDFAG. *T_train* denotes the number of training samples in the target dataset. *n* denotes the feature dimension in the common latent space.

|           | T_train = 5 | T_train = 10 | T_train = 15 | T_train = 20 | T_train = 25 |
| --------- | ----------- | ------------ | ------------ | ------------ | ------------ |
| $n = 50$  | 46.55 s     | 58.60 s      | 73.58 s      | 87.33 s      | 106.91 s     |
| $n = 100$ | 64.22 s     | 87.85 s      | 99.54 s      | 118.44 s     | 147.49 s     |
| $n = 150$ | 80.94 s     | 103.12 s     | 128.87 s     | 156.53 s     | 187.02 s     |
| $n = 200$ | 98.57 s     | 126.63 s     | 159.42 s     | 200.74 s     | 229.20 s     |
| $n = 250$ | 132.59 s    | 153.89 s     | 185.90 s     | 232.15 s     | 241.18 s     |
| $n = 300$ | 161.46 s    | 202.49 s     | 248.43 s     | 291.85 s     | 304.47 s     |
| $n = 350$ | 181.34 s    | 223.77 s     | 274.69 s     | 345.99 s     | 423.55 s     |
| $n = 400$ | 183.73 s    | 236.03 s     | 298.81 s     | 408.99 s     | 478.81 s     |
| $n = 450$ | 187.53 s    | 257.06 s     | 328.63 s     | 415.07 s     | 511.29 s     |
| $n = 500$ | 212.11 s    | 266.83 s     | 338.42 s     | 415.50 s     | 546.41 s     |

Table 5: Average recognition precision (%) comparison of the proposed method CDFAG with state-of-the-art methods. For each method, the best accuracies of various *T_train* values are shown. The best accuracies of each *T_train* are marked in italics.

| Method        | T_train = 5      | T_train = 10     | T_train = 15     | T_train = 20     | T_train = 25     |
| ------------- | ---------------- | ---------------- | ---------------- | ---------------- | ---------------- |
| KEMA [15]     | 46.92 ± 3.26     | 55.75 ± 4.92     | 60.92 ± 2.97     | 63.92 ± 3.37     | 69.42 ± 3.44     |
| SSMA [16]     | 44.33 ± 2.90     | 58.58 ± 1.83     | 63.00 ± 2.38     | 64.92 ± 2.86     | 68.83 ± 1.12     |
| DA [17]       | 41.67 ± 2.52     | 47.83 ± 3.39     | 54.17 ± 1.41     | 55.17 ± 1.52     | 63.67 ± 1.87     |
| WSCDDL [18]   | 45.92 ± 1.26     | 56.08 ± 4.23     | 62.58 ± 2.52     | 66.08 ± 2.03     | 67.42 ± 2.42     |
| Dual [19]     | 35.75 ± 3.23     | 41.83 ± 5.48     | 52.25 ± 2.54     | 58.92 ± 2.22     | 58.25 ± 4.32     |
| No Adaptation | 33.83 ± 3.22     | 46.75 ± 3.53     | 54.17 ± 4.31     | 59.75 ± 3.14     | 66.50 ± 2.01     |
| Baseline      | 30.25 ± 5.41     | 40.33 ± 7.88     | 63.92 ± 2.24     | 66.92 ± 1.37     | 69.25 ± 2.71     |
| *CDFAG*       | *50.75 ± 2.19*   | *62.42 ± 2.59*   | *68.08 ± 3.87*   | *71.58 ± 2.42*   | *75.42 ± 3.13*   |

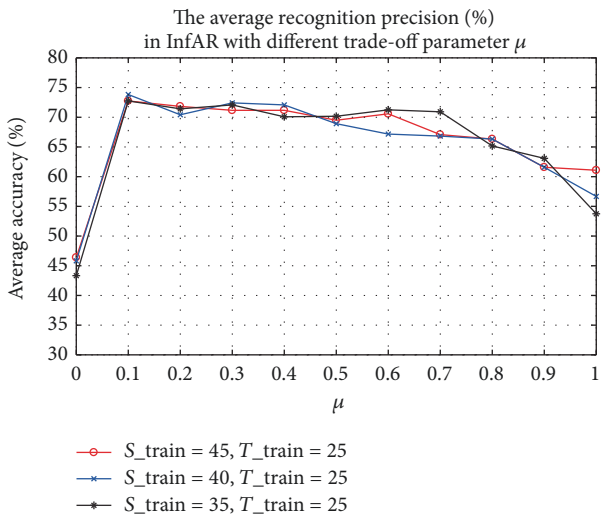

Figure 10: Trade-off parameter $\mu$ and the average recognition precision (%) in InfAR.

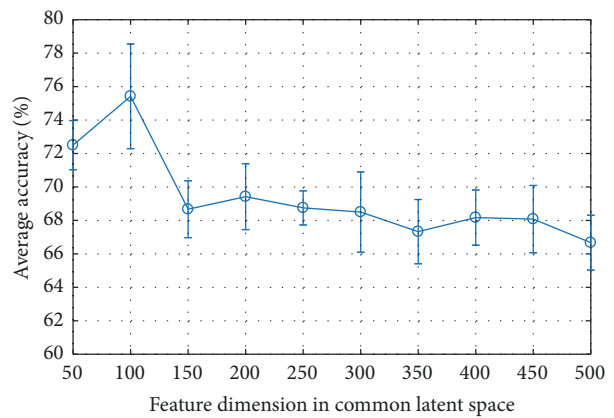

Figure 11: Feature dimension $n$ in latent space and classification accuracy (%).

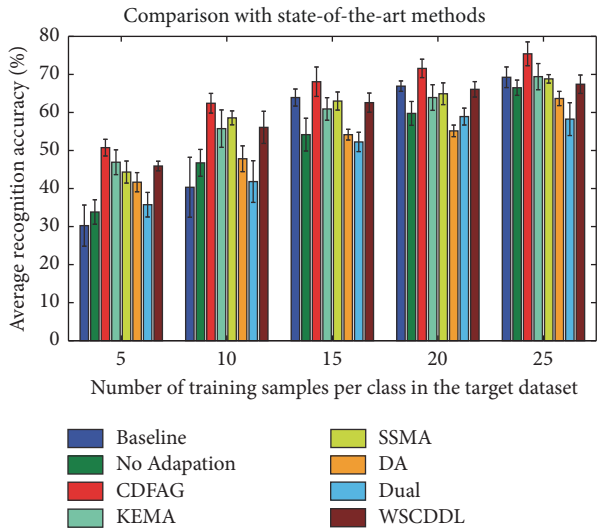

FIGURE 12: Comparison between our results and related works.

We also compare the proposed CDFAG with state-of-the-art deep learning based method. Two-stream CNN is adopted in [20] and achieved average precision 76.66% in InfAR dataset, while the best accuracy of our proposed method is 75.42%. It is evident that our proposed CDFAG can achieve comparable infrared action recognition performance to that of the deep learning based method in [20]. In addition, the proposed CDFAG can still achieve good performance efficiently with fewer labeled training samples in target dataset while the deep learning based method in [20] is time-consuming and needs a large amount of training instances or pretrained models to perform well. Therefore, our proposed CDFAG may be a good visible-to-infrared action recognition framework that makes a good balance between classification accuracy and time efficiency.

## 5. Conclusion and Future Work

In this paper, we propose a novel Cross-Dataset Feature Alignment and Generalization (CDFAG) framework for visible-to-infrared action recognition. The proposed CDFAG is essentially a feature extractor that finds projections from all the source and target domains into a common latent feature space where features of all instances are semantically aligned and generalized with low intraclass diversity and high interclass variance. Promising results are achieved on visible-to-infrared action recognition and cross-dataset recognition tasks, where auxiliary source domain knowledge is effectively transferred to target domain. Compared with several state-of-the-art transfer learning or domain adaptation based methods, our proposed CDFAG offers a more flexible framework and achieves the best performance (AP = 75.42%) in infrared action recognition. In addition, our proposed CDFAG also achieves comparable infrared action recognition performance to deep learning based method. In the future, we will adapt the CDFAG method to other cross-domain action recognition tasks, such as cross-view, cross-dataset, and image-to-video action recognition. Another interesting direction is to modify existing visible light action recognition methods for infrared datasets.

## Conflicts of Interest

The authors declare that there are no conflicts of interest regarding the publication of this paper.

## Acknowledgments

This work is supported by the National Natural Science Foundation of China (no. 61502364), the China Postdoctoral Science Foundation funded project (Grant no. 154906), and the Fundamental Research Funds for the Central Universities (Grant no. 3102016ZY022).

## References

[1] X. Wang, "Intelligent multi-camera video surveillance: A review," *Pattern Recognition Letters*, vol. 34, no. 1, pp. 3–19, 2013.

[2] M. Gori, M. Lippi, M. Maggini, and S. Melacci, "Semantic video labeling by developmental visual agents," *Computer Vision and Image Understanding*, vol. 146, pp. 9–26, 2016.

[3] W. Hu, D. Xie, Z. Fu, W. Zeng, and S. Maybank, "Semantic-based surveillance video retrieval," *IEEE Transactions on Image Processing*, vol. 16, no. 4, pp. 1168–1181, 2007.

[4] W. Zhang, M. L. Smith, L. N. Smith, and A. Farooq, "Gender and gaze gesture recognition for human-computer interaction," *Computer Vision and Image Understanding*, vol. 149, pp. 32–50, 2016.

[5] W. Chen, T. Lao, J. Xia et al., "GameFlow: Narrative Visualization of NBA Basketball Games," *IEEE Transactions on Multimedia*, vol. 18, no. 11, pp. 2247–2256, 2016.

[6] J. K. Aggarwal and M. S. Ryoo, "Human activity analysis: a review," *ACM Computing Surveys*, vol. 43, no. 3, article 16, 2011.

[7] I. Laptev, "On space-time interest points," *International Journal of Computer Vision*, vol. 64, no. 2-3, pp. 107–123, 2005.

[8] H. Wang, A. Kläser, C. Schmid, and C.-L. Liu, "Action recognition by dense trajectories," in *Proceedings of the IEEE Conference on Computer Vision and Pattern Recognition (CVPR '11)*, pp. 3169–3176, June 2011.

[9] H. Bilen, B. Fernando, E. Gavves, A. Vedaldi, and S. Gould, "Dynamic image networks for action recognition," in *Proceedings of the 2016 IEEE Conference on Computer Vision and Pattern Recognition, CVPR 2016*, pp. 3034–3042, usa, July 2016.

[10] C. Schüldt, I. Laptev, and B. Caputo, "Recognizing human actions: a local SVM approach," in *Proceedings of the 17th International Conference on Pattern Recognition (ICPR '04)*, pp. 32–36, August 2004.

[11] H. Kuehne, H. Jhuang, E. Garrote, T. Poggio, and T. Serre, "HMDB: a large video database for human motion recognition," in *Proceedings of the IEEE International Conference on Computer Vision (ICCV '11)*, pp. 2556–2563, November 2011.

[12] K. Soomro, A. R. Zamir, and M. Shah, "UCF101: A dataset of 101 human actions classes from videos in the wild, CoRR abs/1212.0402".

[13] C. Gao, Y. Du, J. Liu, L. Yang, and D. Meng, "A new dataset and evaluation for infrared action recognition," *Communications in Computer and Information Science*, vol. 547, pp. 302–312, 2015.

[14] J. Han and B. Bhanu, "Fusion of color and infrared video for moving human detection," *Pattern Recognition*, vol. 40, no. 6, pp. 1771–1784, 2007.

[15] D. Tuia and G. Camps-Valls, "Kernel manifold alignment for domain adaptation," *PLoS ONE*, vol. 11, no. 2, Article ID e0148655, 2016.

[16] C. Wang and S. Mahadevan, "Heterogeneous domain adaptation using manifold alignment," in *Proceedings of the 22nd International Joint Conference on Artificial Intelligence*, pp. 1541–1546, July 2011.

[17] B. Fernando, A. Habrard, M. Sebban, and T. Tuytelaars, "Unsupervised visual domain adaptation using subspace alignment," in *Proceedings of the 2013 14th IEEE International Conference on Computer Vision, ICCV 2013*, pp. 2960–2967, Australia, December 2013.

[18] F. Zhu and L. Shao, "Weakly-supervised cross-domain dictionary learning for visual recognition," *International Journal of Computer Vision*, vol. 109, no. 1-2, pp. 42–59, 2014.

[19] T. Xu, F. Zhu, E. K. Wong, and Y. Fang, "Dual many-to-one-encoder-based transfer learning for cross-dataset human action recognition," *Image and Vision Computing*, vol. 55, pp. 127–137, 2016.

[20] C. Gao, Y. Du, J. Liu et al., "InfAR dataset: Infrared action recognition at different times," *Neurocomputing*, vol. 212, pp. 36–47, 2016.

[21] X. Yao, J. Han, D. Zhang, and F. Nie, "Revisiting co-saliency detection: a novel approach based on two-stage multi-view spectral rotation co-clustering," *IEEE Transactions on Image Processing*, vol. 26, no. 7, pp. 3196–3209, 2017.

[22] G. Cheng, Z. Li, X. Yao, L. Guo, and Z. Wei, "Remote Sensing Image Scene Classification Using Bag of Convolutional Features," *IEEE Geoscience and Remote Sensing Letters*, vol. 14, no. 10, pp. 1735–1739, 2017.

[23] L. Yang, X. Cai, S. Pan, H. Dai, and D. Mu, "Multi-document summarization based on sentence cluster using non-negative matrix factorization," *Journal of Intelligent & Fuzzy Systems: Applications in Engineering and Technology*, vol. 33, no. 3, pp. 1867–1879, 2017.

[24] C. Yao, J. Han, F. Nie, F. Xiao, and X. Li, "Local regression and global information-embedded dimension reduction, to be published," *IEEE Transactions on Neural Networks and Learning Systems*, 2017.

[25] C. Yao, Y.-F. Liu, B. Jiang, J. Han, and J. Han, "LLE score: a new filter-based unsupervised feature selection method based on nonlinear manifold embedding and its application to image recognition," *IEEE Transactions on Image Processing*, vol. 26, no. 11, pp. 5257–5269, 2017.

[26] S. J. Pan and Q. Yang, "A survey on transfer learning," *IEEE Transactions on Knowledge and Data Engineering*, vol. 22, no. 10, pp. 1345–1359, 2010.

[27] V. M. Patel, R. Gopalan, R. Li, and R. Chellappa, "Visual Domain Adaptation: A survey of recent advances," *IEEE Signal Processing Magazine*, vol. 32, no. 3, pp. 53–69, 2015.

[28] L. Zhang, W. Zuo, and D. Zhang, "LSDT: latent sparse domain transfer learning for visual adaptation," *IEEE Transactions on Image Processing*, vol. 25, no. 3, pp. 1177–1191, 2016.

[29] J. Zhang, J. Yang, and D. Zhang, "Domain class consistency based transfer learning for image classification across domains," *Information Sciences*, vol. 418-419, pp. 242–257, 2017.

[30] L. Zhang and D. Zhang, "Robust visual knowledge transfer via extreme learning machine-based domain adaptation," *IEEE Transactions on Image Processing*, vol. 25, no. 10, pp. 4959–4973, 2016.

[31] J. Ham, D. D. Lee, and L. K. Saul, "Semisupervised alignment of manifolds," in *Proceedings of the 10th International Workshop on Artificial Intelligence and Statistics, AISTATS 2005*, pp. 120–127, brb, January 2005.

[32] D. Tuia, D. Marcos, and G. Camps-Valls, "Multi-temporal and multi-source remote sensing image classification by nonlinear relative normalization," *ISPRS Journal of Photogrammetry and Remote Sensing*, vol. 120, pp. 1–12, 2016.

[33] D. H. Hu, V.-W. Zheng, and Q. Yang, "Cross-domain activity recognition via transfer learning," *Pervasive and Mobile Computing*, vol. 7, no. 3, pp. 344–358, 2011.

[34] A. Farhadi and M. K. Tabrizi, "Learning to Recognize Activities from the Wrong View Point," in *Computer Vision – ECCV 2008*, vol. 5302 of *Lecture Notes in Computer Science*, pp. 154–166, Springer Berlin Heidelberg, Berlin, Heidelberg, 2008.

[35] J. Zheng, Z. Jiang, P. Jonathon Phillips, and R. Chellappa, "Cross-view action recognition via a transferable dictionary pair," in *Proceedings of the 2012 23rd British Machine Vision Conference, BMVC 2012*, gbr, September 2012.

[36] Z. Zhang, C. Wang, B. Xiao, W. Zhou, S. Liu, and C. Shi, "Cross-view action recognition via a continuous virtual path," in *Proceedings of the 26th IEEE Conference on Computer Vision and Pattern Recognition (CVPR '13)*, pp. 2690–2697, June 2013.

[37] X. Wu, H. Wang, C. Liu, and Y. Jia, "Cross-view action recognition over heterogeneous feature spaces," *IEEE Transactions on Image Processing*, vol. 24, no. 11, pp. 4096–4108, 2015.

[38] W. Sui, X. Wu, Y. Feng, and Y. Jia, "Heterogeneous discriminant analysis for cross-view action recognition," *Neurocomputing*, vol. 191, pp. 286–295, 2016.

[39] C. Zu and D. Zhang, "Canonical sparse cross-view correlation analysis," *Neurocomputing*, vol. 191, pp. 263–272, 2016.

[40] W. Bian, D. Tao, and Y. Rui, "Cross-domain human action recognition," *IEEE Transactions on Systems, Man, and Cybernetics, Part B: Cybernetics*, vol. 42, no. 2, pp. 298–307, 2012.

[41] N. C. Tang, Y.-Y. Lin, J.-H. Hua, S.-E. Wei, M.-F. Weng, and H. M. Liao, "Robust action recognition via borrowing information across video modalities," *IEEE Transactions on Image Processing*, vol. 24, no. 2, pp. 709–723, 2015.

[42] F. Liu, X. Xu, S. Qiu, C. Qing, and D. Tao, "Simple to complex transfer learning for action recognition," *IEEE Transactions on Image Processing*, vol. 25, no. 2, pp. 949–960, 2016.

[43] M. Kan, S. Shan, and X. Chen, "Bi-shifting auto-encoder for unsupervised domain adaptation," in *Proceedings of the 15th IEEE International Conference on Computer Vision, ICCV 2015*, pp. 3846–3854, Chile, December 2015.

[44] J. Han and B. Bhanu, "Human Activity Recognition in Thermal Infrared Imagery," in *Proceedings of the 2005 IEEE Computer Society Conference on Computer Vision and Pattern Recognition (CVPR'05) - Workshops*, p. 17, San Diego, CA, USA.

[45] H. Eum, J. Lee, C. Yoon, and M. Park, "Human action recognition for night vision using temporal templates with infrared thermal camera," in *Proceedings of the 2013 10th International Conference on Ubiquitous Robots and Ambient Intelligence, URAI 2013*, pp. 617–621, Republic of Korea, November 2013.

[46] Y. Zhu and G. Guo, "A study on visible to infrared action recognition," *IEEE Signal Processing Letters*, vol. 20, no. 9, pp. 897–900, 2013.

[47] J. Yang, R. Yan, and A. G. Hauptmann, "Cross-domain video concept detection using adaptive SVMs," in *Proceedings of the 15th ACM International Conference on Multimedia (MM '07)*, pp. 188–197, September 2007.

[48] H. Wang and C. Schmid, "Action recognition with improved trajectories," in *Proceedings of the 14th IEEE International Conference on Computer Vision (ICCV '13)*, pp. 3551–3558, Sydney, Australia, December 2013.

[49] J. Wang, J. Yang, K. Yu, F. Lv, T. Huang, and Y. Gong, "Locality-constrained linear coding for image classification," in *Proceedings of the IEEE Computer Society Conference on Computer Vision and Pattern Recognition (CVPR '10)*, pp. 3360–3367, IEEE, June 2010.

[50] L. I. Smith, "A tutorial on principal components analysis," *Information Fusion*, vol. 51, no. 3, pp. 219–226, 2002.

[51] B. L. V. D. Waerden, *The Mathematical Gazette*, Prentice-Hall Internations, Inc, 1971.

[52] S. Yan, D. Xu, B. Zhang, H. Zhang, Q. Yang, and S. Lin, "Graph embedding and extensions: a general framework for dimensionality reduction," *IEEE Transactions on Pattern Analysis and Machine Intelligence*, vol. 29, no. 1, pp. 40–51, 2007.

[53] Y. Fang, J. Xie, G. Dai et al., "3D deep shape descriptor," in *Proceedings of the IEEE Conference on Computer Vision and Pattern Recognition, CVPR 2015*, pp. 2319–2328, USA, June 2015.

[54] A. Coates, H. Lee, and A. Y. Ng, "An analysis of single-layer networks in unsupervised feature learning," *Journal of Machine Learning Research*, vol. 15, pp. 215–223, 2011.

[55] K. Nogueira, O. A. B. Penatti, and J. A. dos Santos, "Towards better exploiting convolutional neural networks for remote sensing scene classification," *Pattern Recognition*, vol. 61, pp. 539–556, 2017.

[56] G. Cheng, C. Yang, X. Yao, L. Guo, and J. Han, "When Deep Learning Meets Metric Learning: Remote Sensing Image Scene Classification via Learning Discriminative CNNs," *IEEE Transactions on Geoscience and Remote Sensing*, pp. 1–11.

[57] J. M. Chaquet, E. J. Carmona, and A. Fernández-Caballero, "A survey of video datasets for human action and activity recognition," *Computer Vision and Image Understanding*, vol. 117, no. 6, pp. 633–659, 2013.

[58] C. Chang and C. Lin, "LIBSVM: a Library for support vector machines," *ACM Transactions on Intelligent Systems and Technology*, vol. 2, no. 3, article 27, 2011.

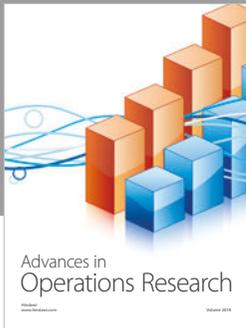

Advances in
**Operations Research**

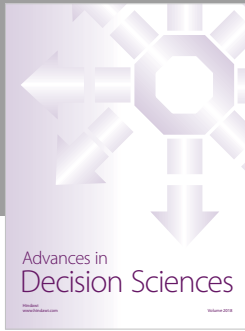

Advances in
**Decision Sciences**

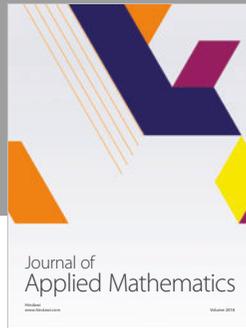

Journal of
**Applied Mathematics**

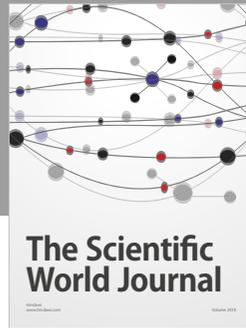

**The Scientific
World Journal**

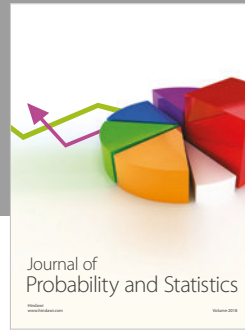

Journal of
**Probability and Statistics**

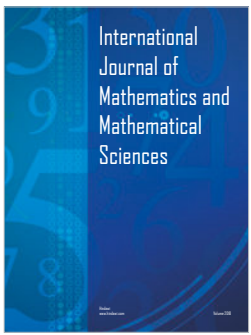

**International
Journal of
Mathematics and
Mathematical
Sciences**

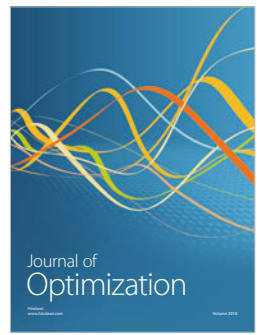

Journal of
**Optimization**

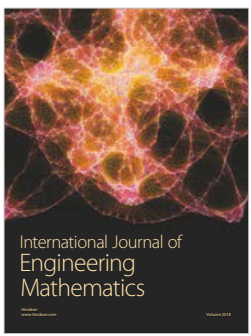

**International Journal of
Engineering
Mathematics**

Submit your manuscripts at
www.hindawi.com

**Hindawi**

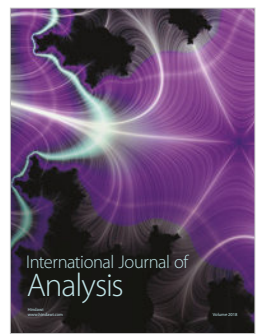

**International Journal of
Analysis**

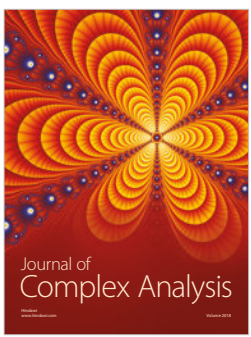

Journal of
**Complex Analysis**

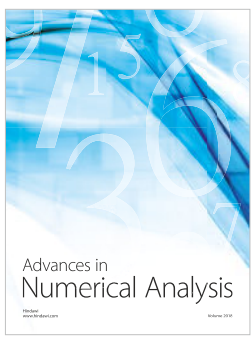

Advances in
**Numerical Analysis**

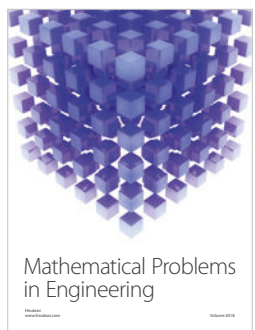

**Mathematical Problems
in Engineering**

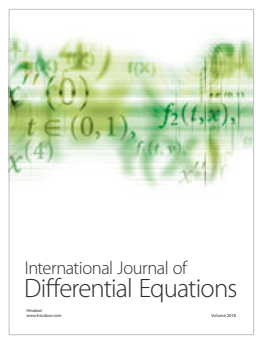

International Journal of
**Differential Equations**

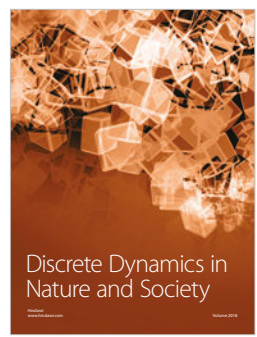

**Discrete Dynamics in
Nature and Society**

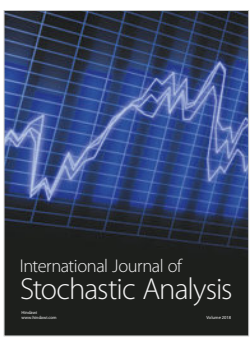

International Journal of
**Stochastic Analysis**

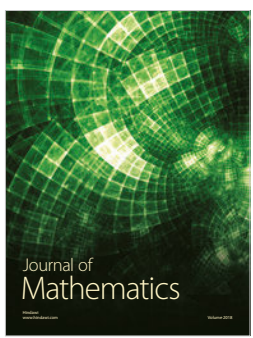

Journal of
**Mathematics**

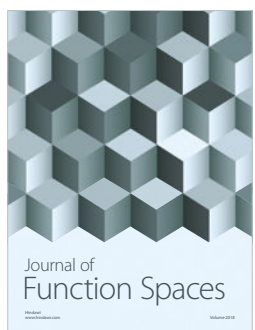

Journal of
**Function Spaces**

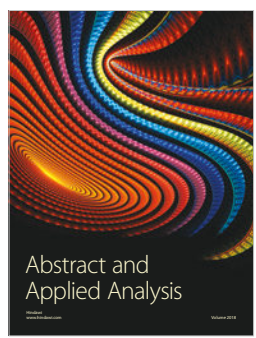

**Abstract and
Applied Analysis**

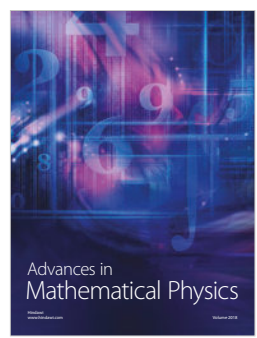

Advances in
**Mathematical Physics**